\documentclass[final]{cvpr}

\usepackage{times}
\usepackage{epsfig}
\usepackage{graphicx}
\usepackage{amsmath}
\usepackage{amssymb}

\usepackage{url}            
\usepackage{booktabs}       
\usepackage{amsfonts}       
\usepackage{nicefrac}       
\usepackage{microtype}      

\usepackage{times}
\usepackage{epsfig}
\usepackage{graphicx}
\usepackage{amsmath}
\usepackage{amssymb}
\usepackage{multirow}
\usepackage{color}
\usepackage{algorithmic}
\usepackage{algorithm}
\usepackage{booktabs}
\usepackage{url}
\usepackage[table]{xcolor}
\usepackage{colortbl}
\usepackage{bm}
\usepackage{subcaption}
\usepackage{makecell}
\usepackage{pifont}

\usepackage{threeparttable}

\newcommand{\cmark}{\ding{51}}%
\newcommand{\xmark}{\ding{55}}%

\newcommand{\more}[1]{\small{\textcolor[RGB]{57,181,74}{#1}}}

\usepackage[pagebackref=true,breaklinks=true,colorlinks,bookmarks=false]{hyperref}

\begin{document}

\title{Distribution Alignment: A Unified Framework for Long-tail Visual Recognition}

\author{Songyang Zhang$^{1,3,5,}$\thanks{This work was done when Songyang Zhang was a research intern at Megvii Technology. This work was supported by Shanghai NSF Grant (No. 18ZR1425100), National Key R\&D Program of China (No. 2017YFA0700800), and Beijing Academy of Artificial Intelligence (BAAI). Code is available: \href{https://github.com/Megvii-BaseDetection/DisAlign}{https://github.com/Megvii-BaseDetection/DisAlign}}\quad Zeming Li$^{2}$\quad Shipeng Yan$^{1}$\quad
Xuming He$^{1,4}$\quad Jian Sun$^{2}$\\
$^1$ShanghaiTech University\quad $^2$Megvii Technology\quad $^3$ University of Chinese Academy of Sciences\\ $^4$Shanghai Engineering Research Center of Intelligent Vision and Imaging\\$^5$Shanghai Institute of Microsystem and Information Technology, Chinese Academy of Sciences \\
{\tt\small \{zhangsy1, yanshp, hexm\}@shanghaitech.edu.cn, \{lizeming,sunjian\}@megvii.com}\\
}

\maketitle

\begin{abstract}
Despite the recent success of deep neural networks, it remains challenging to effectively model the long-tail class distribution in visual recognition tasks. To address this problem, we first investigate the performance bottleneck of the two-stage learning framework via ablative study. Motivated by our discovery, we propose a unified distribution alignment strategy for long-tail visual recognition. Specifically, we develop an adaptive calibration function that enables us to adjust the classification scores for each data point. We then introduce a generalized re-weight method in the two-stage learning to balance the class prior, which provides a flexible and unified solution to diverse scenarios in visual recognition tasks. We validate our method by extensive experiments on four tasks, including image classification, semantic segmentation, object detection, and instance segmentation.
Our approach achieves the state-of-the-art results across all four recognition tasks with a simple and unified framework. 
\end{abstract}

\vspace{-3mm}
\section{Introduction}

While deep convolutional networks have achieved great successes in many vision tasks, it usually requires a large number of training examples for each visual category. More importantly, prior research mostly focuses on learning from a \textit{balanced dataset}~\cite{krizhevsky2012imagenet}, where different object classes are approximately evenly distributed. However, for large-scale vision recognition tasks, partially due to the non-uniform distribution of natural object classes and varying annotation costs, we typically learn from datasets with a \textit{long-tail} class label distribution. In such scenarios, the number of training instances per class varies significantly, from as few as one example for tail classes to hundreds or thousands for head classes~\cite{zhou2017places, liu2019large, gupta2019lvis, zhu2014capturing, zhou2017scene,van2018inaturalist}.

\begin{figure}
    \centering
    \includegraphics[width=0.9\linewidth]{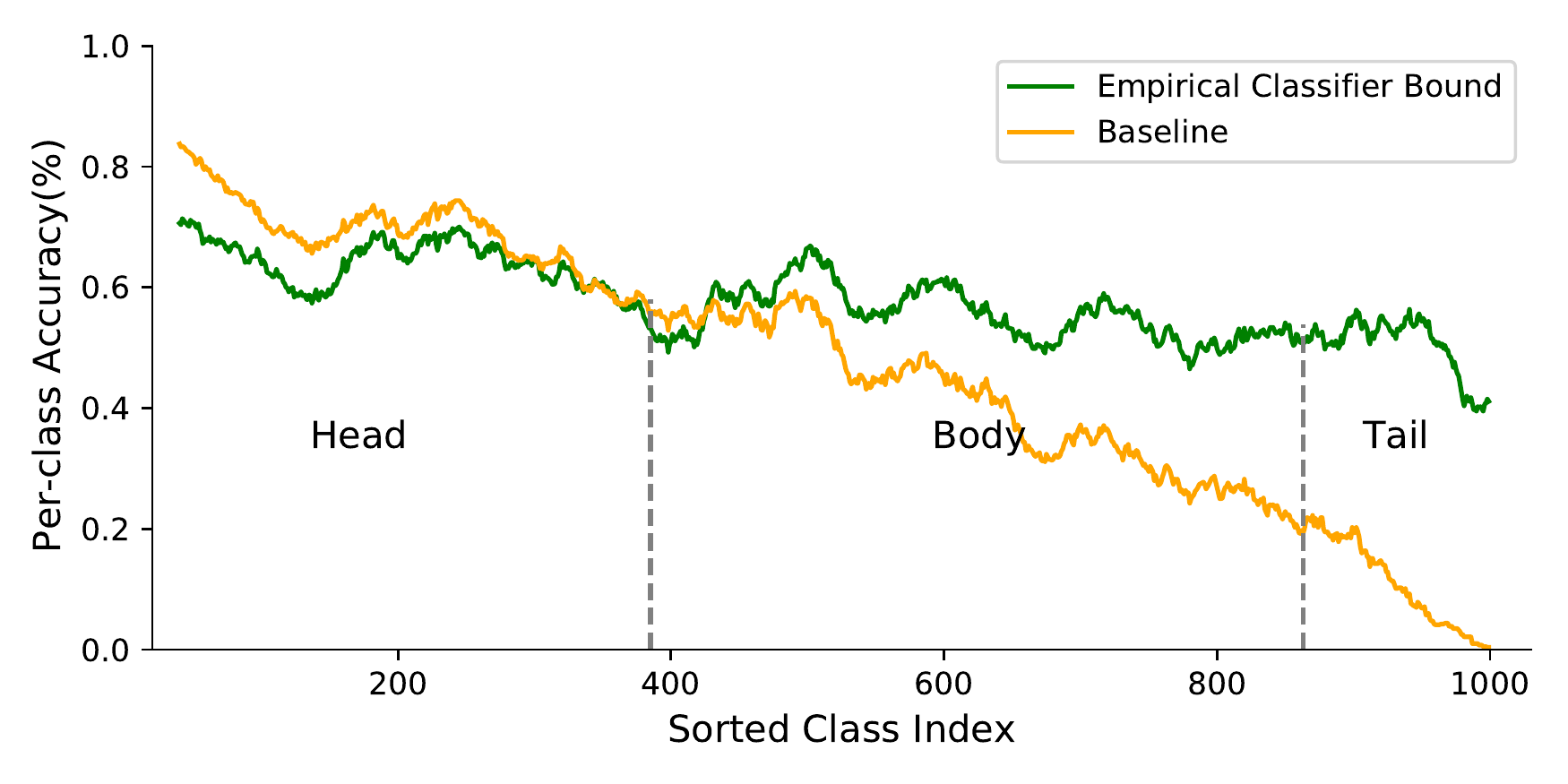}
    \caption{Per-class performance of the two-stage learning \textit{baseline} and our \textit{empirical classification bound} on ImageNet-LT val split. Two methods share the same representation while our bound setting retrains the classifier head with the balanced full dataset.}\vspace{-3mm}
    \label{fig:ad}
\end{figure}

The intrinsic long-tail property of our visual data introduces a multitude of challenges for recognition in the wild~\cite{bengio2015sharing}, as a deep network model has to simultaneously cope with imbalanced annotations among the head and medium-sized classes, and few-shot learning in the tail classes. A naively learned model would be largely dominated by those few head classes while its performance is much degraded for many other tail classes.

Early works on re-balancing data distribution focus on learning one-stage models, which achieve limited successes due to lack of principled design in their strategies~\cite{buda2018systematic,shen2016relay,cao2019learning,cui2019class,liu2019large,wang2017learning}. More recent efforts aim to improve the long-tail prediction by decoupling the representation learning and classifier head learning~\cite{kang2019decoupling, menon2020long, tang2020long, wang2020devil,li2020overcoming}. However, such a two-stage strategy typically relies on heuristic design to adjust the decision boundary of the initially learned classifier head, which often requires tedious hyper-parameter tuning in practice. This severely limits its capacity to resolve the mismatch between \textit{imbalanced training data distribution} and \textit{balanced evaluation metrics}.

In this work, we first perform an ablative analysis on the two-stage learning strategy to shed light on its performance bottleneck. Specifically, our study estimates an `ideal' classification accuracy using a balanced dataset to retrain the classifier head while keeping the first-stage representation fixed. Interestingly, as shown in Fig.~\ref{fig:ad}, we find a substantial gap between this ideal performance and the baseline network, which indicates that the first-stage learning with unbalanced data provides a good representation, but there is a large room for improvement in the second stage due to the \textit{biased decision boundary} (See Sec.~\ref{subsec:overview} for details).

Based on those findings, we propose a simple and yet effective two-stage learning scheme for long-tail visual recognition problems. Our approach focuses on improving the second-stage training of the classifier after learning a feature representation in a standard manner. To this end, we develop a \textit{unified} distribution alignment strategy to calibrate the classifier output via matching it to a reference distribution of classes that favors the balanced prediction. Such an alignment strategy enables us to exploit the class prior and data input in a principled manner for learning class decision boundary, which eliminates the needs for tedious hyper-parameter tuning and can be easily applied to various visual recognition tasks.

Specifically, we develop a light-weight distribution alignment module for calibrating classification scores, which consists of two main components. In the first component, we introduce an adaptive calibration function that equips the class scores with an input-dependent, learnable magnitude and margin. This allows us to achieve a flexible and confidence-aware distribution alignment for each data point. Our second component explicitly incorporates a balanced class prior by employing a generalized re-weight design for the reference class distribution, which provides a unified strategy to cope with diverse scenarios of label imbalance in different visual recognition tasks.

We extensively validate our model on four typical visual recognition tasks, including image classification on three benchmarks (ImageNet-LT~\cite{liu2019large}, iNaturalist~\cite{van2018inaturalist} and Places365-LT~\cite{liu2019large}), semantic segmentation on ADE20k dataset~\cite{zhou2017scene}, object detection and instance segmentation on LVIS dataset~\cite{gupta2019lvis}. The empirical results and ablative study show our method consistently outperforms the state-of-the-art approaches on all the benchmarks. 
To summarize, the main contributions of our works are three-folds:
\begin{itemize}
	\vspace{-0.3em}
	\item We conduct an empirical study to investigate the performance bottleneck of long-tail recognition and reveal a critical gap caused by biased decision boundary.
	\vspace{-0.3em}
	\item We develop a simple and effective distribution alignment strategy with a generalized re-weight method, which can be easily optimized for various long-tail recognition tasks without whistles and bells.
	\vspace{-0.3em}
	\item Our models outperform previous work with a large margin and achieve state-of-the-art performance on long-tail image classification, semantic segmentation, object detection, and instance segmentation.
	\end{itemize}

\section{Related Works}

%

\paragraph{One-stage Imbalance Learning}
To alleviate the adverse effect of the long-tail class distribution in visual recognition, prior work have extensively studied the one-stage methods, which either leverage the re-balancing ideas or explore knowledge transfer from head categories.
The basic idea of resample-based methods is to over-sample the minority categories~\cite{chawla2002smote,han2005borderline} or to under-sample the frequent categories in the training process~\cite{drummond2003c4, buda2018systematic}. Class-aware sampling~\cite{shen2016relay} proposes to choose samples of each category with equal probabilities, which is widely used in vision tasks~\cite{liu2019large, gao2018solution}. Repeat factor sampling ~\cite{mahajan2018exploring} is a smoothed sampling method conducting repeated sampling for tail categories, which demonstrates its efficacy in instance segmentation~\cite{gupta2019lvis}. In addition, \cite{wang2019data} proposes to increase the sampling rate for categories with low performance after each training epoch and balances the feature learning for under-privileged categories. 

An alternative strategy is to re-weight the loss function in training. Class-level methods typically re-weight the standard loss with category-specific coefficients correlated with the sample distributions~\cite{huang2016learning,cui2019class,cao2019learning,khan2017cost,khan2019striking, tan2020equalization}. Sample-level methods~\cite{lin2017focal,ren2018learning} try to introduce a more fine-grained control of loss for imbalanced learning. 
Other work aim to enhance the representation or classifier head of tail categories by transferring knowledge from the head classes~\cite{wang2017learning, wang2018low,liu2019large, zhong2019unequal,chu2020feature,wu2020solving,wu2020self}. 
Nevertheless, these methods require designing a task specific network module or structure, which is usually non-trivial to be generalized to different vision tasks.

\vspace{-3mm}
\paragraph{Two-stage Imbalance Learning}
 More recent efforts aims to improve the long-tail prediction by decoupling the learning of representation and classifier head.
 Decouple~\cite{kang2019decoupling} proposes an instance-balanced sampling scheme, which generates more generalizable representations and achieves strong performance after properly re-balancing the classifier heads. The similar idea is adopted in~\cite{wang2020devil,wang2020few,li2020overcoming}, which develop effective strategies for long-tail object detection tasks. \cite{menon2020long, tang2020long} improve the two-stage ideas by introducing a post-process to adjust the prediction score. However, such a two-stage strategy typically relies on heuristic design in order to adjust the decision boundary of initially learned classifiers and requires tedious hyper-parameter tuning in practice.

\vspace{-3mm}
\paragraph{Visual Recognition Tasks} 
Visual recognition community has witnessed significant progress with deep convolutional networks in recent years. 
In this study, we focus on four types of visual tasks, including image classification, object detection, semantic and instance segmentation, which have been actively studied in a large amount of prior work.  
For object detection,  we consider the typical deep network architecture used in the R-CNN series method~\cite{girshick2014rich, girshick2015fast, ren2015faster}, which detects objects based on the region proposals. For instance segmentation, we take the 
 Mask R-CNN~\cite{he2017mask} as our example, which extends the Faster R-CNN\cite{ren2015faster} by adding a branch for predicting the object masks in parallel with the existing branch for bounding box recognition.
For the pixel-wise task, semantic segmentation, we use the FCN-based methods~\cite{shelhamer2017fully}  
and the widely-adopted encoder-decoder structures~\cite{deeplabv3plus2018,chen2018deeplab,chen2017rethinking}. 
Despite those specific choices, we note that our strategy can be easily extended to other types of deep network methods for those visual recognition tasks. 

\section{Our Approach}

Our goal is to address the problem of large-scale long-tail visual recognition, which typically has a large number of classes and severe class imbalance in its training data. To this end, we adopt a two-stage learning framework that first learns a feature representation and a classifier head from the unbalanced data, followed by a calibration stage that adjusts the classification scores. Inspired by our ablative study on existing two-stage methods, we propose a principled calibration method that aligns the model prediction with a reference class distribution favoring the balanced evaluation metrics. Our distribution alignment strategy is simple and yet effective, enabling us to tackle different types of large-scale long-tail visual recognition tasks in a unified framework.

Below we start with a brief introduction to the long-tail classification and an empirical study of two-stage methods in Sec.\ref{subsec:overview}. We then describe our proposed distribution alignment strategy in Sec.\ref{subsec:alignment}. Finally, we present a comparison with previous methods from the distribution match perspective in Sec.\ref{subsec:distribution_mismatch}.

\subsection{Problem Setting and Empirical Study}\label{subsec:overview}

We now introduce the problem setting of long-tail classification and review the two-stage learning framework for deep networks. Subsequently, we perform an empirical ablative study on a large-scale image classification task, which motivates our proposed approach. 

\paragraph{Problem Definition}

The task of long-tail recognition aims to learn a classification model from a training dataset with long-tail class distribution. 
Formally, we denote the input as $\mathbf{I}$, and the target label space as $\mathcal{C}=\{c_1,\cdots,c_K\}$, where $K$ is the number of classes. The classification model $\mathcal{M}$ defines a mapping from the input to the label space: $y = \mathcal{M}(\mathbf{I};\Theta)$, where $y\in\mathcal{C}$ and $\Theta$ are its parameters. Our goal is to learn the model parameter from an imbalanced training dataset 
$\mathcal{D}_{tr}$ so that $\mathcal{M}$ achieves optimal performance on an evaluation dataset $\mathcal{D}_{eval}$ with respect to certain balanced metrics (\textit{e.g}., mean accuracy).  

In the two-stage framework, we typically consider a deep network model $\mathcal{M}$ with two main components: a feature extractor network $f(\cdot)$ and a classifier head $h(\cdot)$. The feature extractor $f$ first extracts an input representation $\mathbf{x}$, which is then fed into the classifier head $h$ to compute class prediction scores $\mathbf{z}$ as follows:
\begin{align}
\mathbf{x} = f(\mathbf{I}, \theta_{f})\in\mathbb{R}^{d}, \quad
\mathbf{z} = h(\mathbf{x},\theta_{h})\in\mathbb{R}^K
\end{align}
where $\theta_{f}$ and $\theta_{h}$ are the parameter of $f(\cdot)$ and $h(\cdot)$, respectively. Here $\mathbf{z}=\{z_1,\cdots,z_K\}$ indicate the class prediction scores for $K$ classes and the model predicts the class label by taking $y=\arg\max{(\mathbf{z})}$. 

In this work, we instantiate the classifier head $h$ as a linear classifier or a cosine similarity classifier as follows:
\begin{align}
\text{Linear}:\quad &z_j=\mathbf{w}_j^\intercal\mathbf{x}\\
\text{Cosine Similarity}:\quad &z_j=s\cdot\frac{\mathbf{w}_j^\intercal\mathbf{x}}{||\mathbf{w}_j||||\mathbf{x}||}
\end{align}
where $\mathbf{w}_j\in\mathbb{R}^d$ is the parameter of $j$-th class and the $s$ is a scale factor as in \cite{qi2018low}. 
We note that the above formulation can be instantiated for multiple visual recognition tasks by changing the input $\mathbf{I}$: e.g., an image for image classification, an image with a pixel location for semantic segmentation, or an image with a bounding box for object detection. 

\begin{figure}[t]
	\centering
	\includegraphics[width=0.9\linewidth]{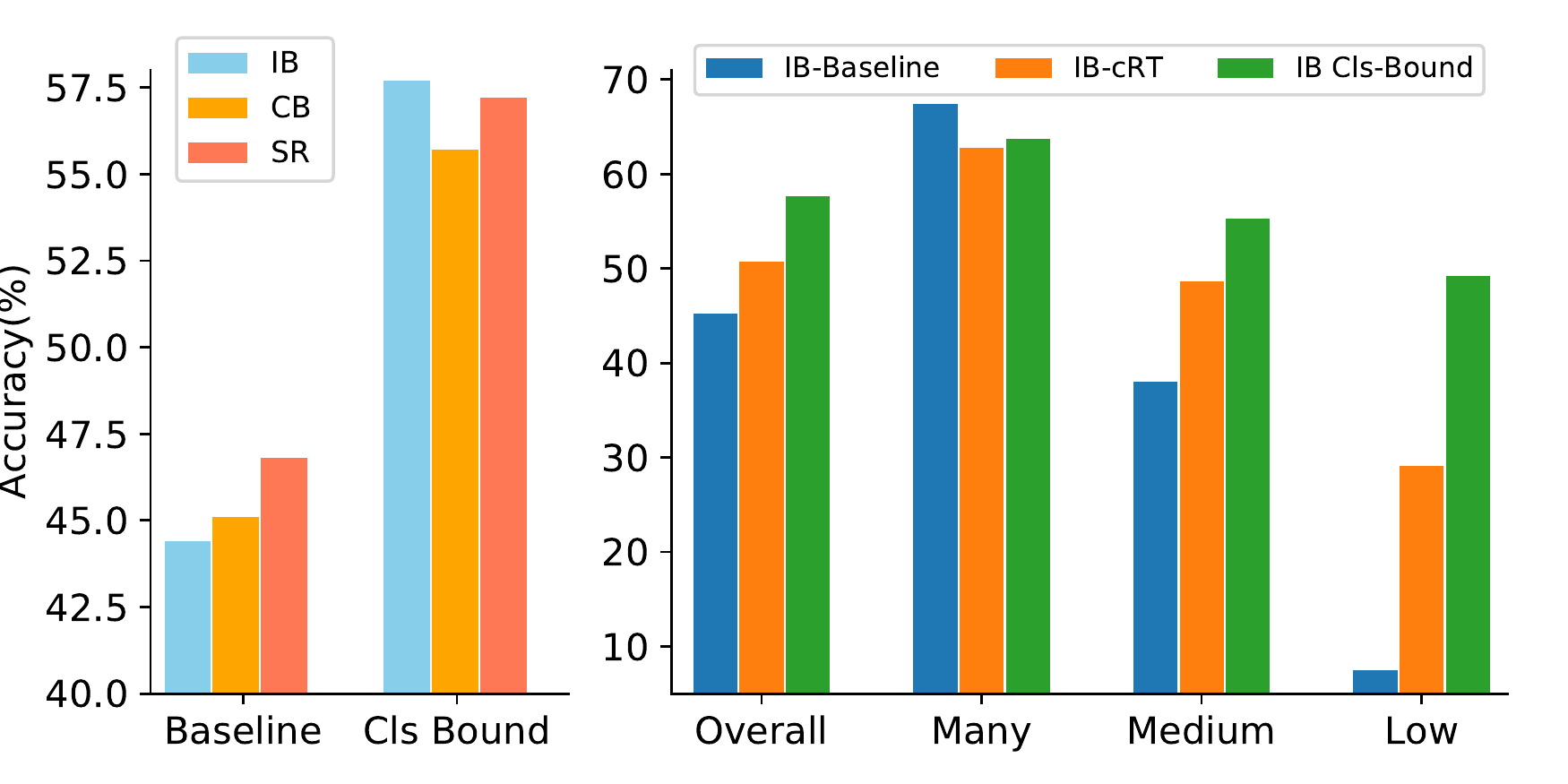}
	\caption{\textbf{Empirical analysis of the performance bottleneck.} \textit{Left}: Baseline vs. ideal performance for representations learned with different sampling strategy. \textit{Right}: Comparison of prior arts and ideal performance for the classifier head calibration. \textbf{Cls-Bound}: ideal performance bound. \textbf{IB}: instance-balanced sampling. \textbf{CB}: class-balanced sampling. \textbf{SR}: square-root sampling.}
	\label{fig:upperbound}
\end{figure}

\paragraph{Empirical Analysis on Performance Bound}

The two-stage learning method tackles the long-tail classification by decoupling the representation and the classifier head learning~\cite{kang2019decoupling}. Specifically, it first learns the feature extractor $f$ and classifier head $h$ jointly, and then with the representation fixed, re-learns the classifier head with a class balancing strategy. 
While such design achieves certain success, an interesting question to ask is \textit{which model component(s) impose a bottleneck on its balanced performance}.  
In the following, we attempt to address the question by exploiting the full set of the ImageNet dataset. Particularly, we follow the decoupling idea to conduct a series of ablative studies on two model components under an `ideal' balanced setting.  

We first investigate whether the feature representation learned on the imbalanced dataset is restrictive for the balanced performance.
To this end, we start from learning the feature extractor on the imbalanced ImageNet-LT training set with several re-balancing strategies (\textit{e.g.} instance-balanced, class-balanced, or square-root sampling). We then keep the representation fixed and re-train the classifier head with the ideal {balanced} ImageNet train set (excluding ImageNet-LT val set). Our results are shown in the left panel of Fig.~\ref{fig:upperbound}, which indicate that \textit{the first stage produces a strong feature representation that can potentially lead to large performance gain and the instance-based sampling achieves better overall results} (cf. \cite{kang2019decoupling}).

Moreover,  we conduct an empirical study on the effectiveness of the recent decoupling method (\textit{e.g}. cRT~\cite{kang2019decoupling}) compared with the above 'ideal' classifier head learning. The right panel of Fig.~\ref{fig:upperbound} shows that there remains \textit{a large performance gap} between the existing methods and the upper-bound.
Those empirical results indicate that \textit{the biased decision boundary in the feature space seems to be the performance bottleneck of the existing long-tail methods}. Consequently, a better strategy to address this problem would further improve the two-stage learning for the long-tail classification.  

\subsection{Distribution Alignment}\label{subsec:alignment}
To tackle the aforementioned issue, we now introduce a \textit{unified} distribution alignment strategy to calibrate the classifier output via matching it to a reference distribution of classes that favors the balanced prediction. In this work, we adopt a two-stage learning scheme for all visual recognition tasks, which consists of {a joint learning stage} and {a distribution calibration stage} as follows. 

\noindent\textit{1) Joint Learning Stage.} The feature extractor $f(\cdot)$ and original classifier head (denoted as $h_o(\cdot)$ for clarity) are jointly learned on imbalanced $\mathcal{D}_{tr}$ with instance-balanced strategy in the first stage, where the original $h_o(\cdot)$ is severely biased due to the imbalanced data distribution. 

\noindent\textit{2) Distribution Calibration Stage.} For the second stage, the parameters of $f(\cdot)$ are frozen and we only focus on the classifier head to adjust the decision boundary. To this end,
we introduce an \textit{adaptive calibration function} (in Sec.~\ref{subsubsec:adaptive_calibration}) and a \textit{distribution alignment strategy with generalized re-weighting} (in Sec.~\ref{subsubsec:generalized_reweight}) to calibrate the class scores. 


\subsubsection{Adaptive Calibration Function}\label{subsubsec:adaptive_calibration}


 \quad To learn the classifier head $h(\cdot)$ in the second stage, we propose an adaptive calibration strategy that fuses the original classifier head $h_o(\cdot)$ (parameters of $h_o(\cdot)$ are frozen) and a learned class prior in an input-dependent manner. As shown below, unlike previous work (\textit{e.g}. cRT\cite{kang2019decoupling}), our design does not require a re-training of the classifier head from scratch and has much fewer free parameters. This enables us to reduce the adverse impact from the limited training data of the tail categories. Moreover, we introduce a flexible fusion mechanism capable of controlling the magnitude of calibration based on input features.

Specifically, denote the class scores  from $h_o(\cdot)$ as $\mathbf{z}^o=[z^o_1,\cdots,z^o_K]$, we first introduce a class-specific linear transform to adjust the score as follows:
\begin{align}
s_j = \alpha_j\cdot z^o_j + \beta_j, \quad \forall j\in\mathcal{C}
\end{align}
where $\alpha_j$ and $\beta_j$ are the calibration parameters for each class, which will be learned from data. As mentioned above, we then define a confidence score function $\sigma(\mathbf{x})$ to adaptively combine the original and the transformed class scores:
\begin{align}
\hat{z}_j &=  \sigma({\mathbf{x}})\cdot s_j + (1-\sigma({\mathbf{x}}))\cdot z^o_j\\
&= (1+\sigma({\mathbf{x}})\alpha_j)\cdot z^o_j + \sigma({\mathbf{x}})\cdot \beta_j \label{eq:final}
\end{align}
where the confidence score has a form of $g(\mathbf{v}^\intercal\mathbf{x})$, which is implemented as a linear layer followed by a non-linear activation function (\textit{e.g.}, sigmoid function) for all input $\mathbf{x}$. The confidence $\sigma(\mathbf{x})$ controls how much calibration is needed for a specific input $\mathbf{x}$. 
Given the calibrated class scores, we finally define a prediction distribution for our model with the Softmax function:
\begin{align}
p_m(y=j|\mathbf{x})=\frac{\exp(\hat{z}_j)}{\sum_{k=1}^C\exp(\hat{z}_k)}.
\end{align}

\subsubsection{Alignment with Generalized Re-weighting}\label{subsubsec:generalized_reweight}

\begin{figure}[t]
	\centering
	\includegraphics[width=0.85\linewidth]{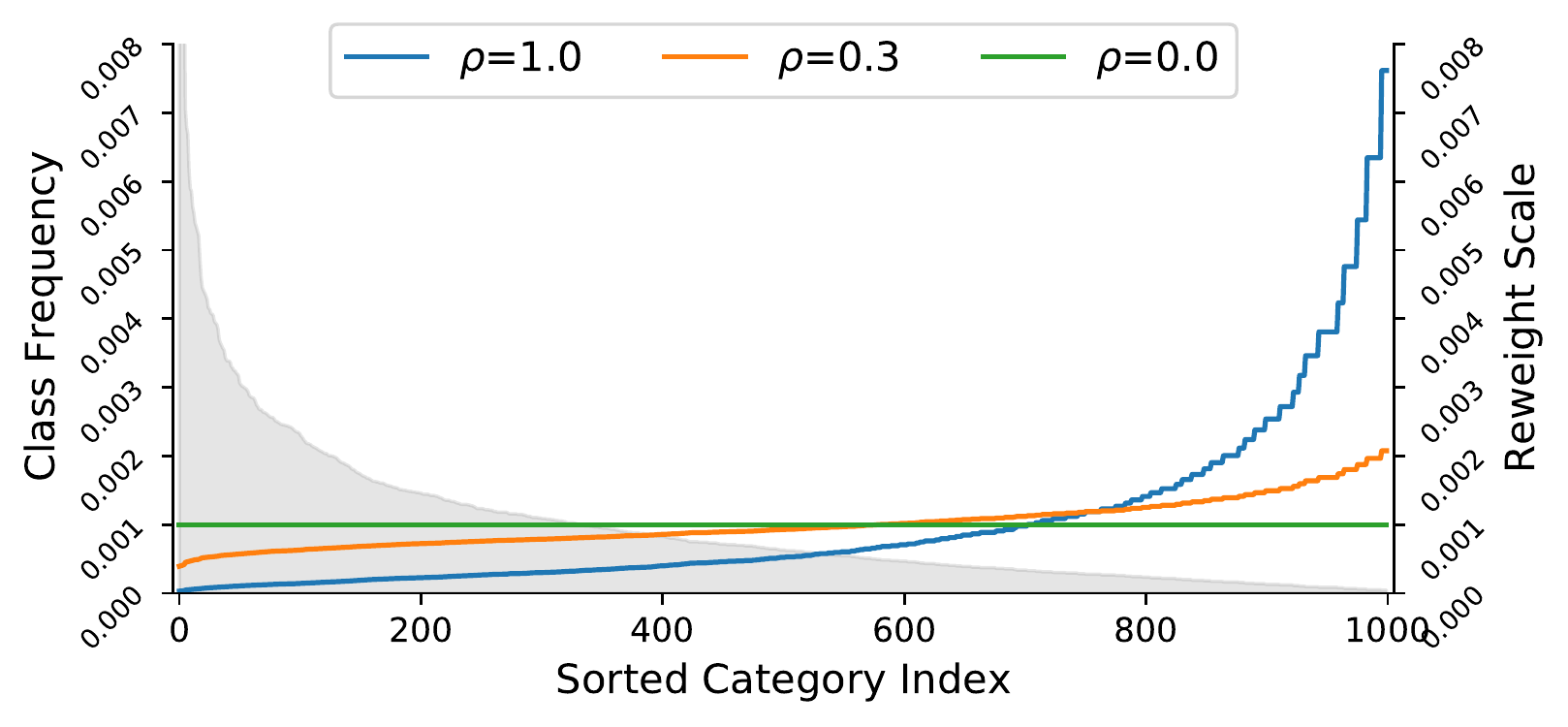}
	\caption{The category frequency $r_i$ is plotted in gray color and the right axis denotes the re-weight coefficient at different scale $\rho$ on ImageNet-LT dataset.}
	\label{fig:exp_scale_illustration}
\end{figure}

\begin{table*}[h!]
	\centering
	\resizebox{0.85\textwidth}{!}{
		\begin{tabular}{l|c|c|ccc|c|ccc}
			\toprule
			\multirow{2}{*}{\textbf{Method}}  & \multirow{2}{*}{\textbf{Align Type}} & \multicolumn{4}{c|}{\textbf{Top-1 Accuracy@R-50}} & \multicolumn{4}{c}{\textbf{Top-1 Accuracy@X-50}}  \\
			\cmidrule{3-10}
			&             &\textbf{Average} & \textbf{Many } & \textbf{Medium  } & \textbf{Few  } &\textbf{Average} & \textbf{Many  } & \textbf{Medium  } & \textbf{Few  } \\
			\midrule
			Baseline\cite{kang2019decoupling}    & -           & 41.6	           & 64.0           &    33.8         &  5.8           & 44.4	          & 65.9           &       37.5     &  7.7 \\
			Baseline$^*$                   & -               & 48.4  & 68.4   &  41.7    & 15.2                 & 49.2   & 68.9   &  42.8  & 15.6\\
			\midrule
			NCM\cite{kang2019decoupling}          & \multirow{4}{*}{ Hand-Craft }  & 44.3             & 53.1           &    42.3         &  26.5         & 47.3         & 56.6          &        45.3    &   28.1\\
			$\tau$-Norm\cite{kang2019decoupling}  & & 46.7             & 56.6           &    44.2         &  27.4         & 49.4         & 59.1          &        46.9    &   30.7    \\
			Logit Adjust{(post)}\cite{menon2020long}                   &   & 50.4             & -              &      -          &   -           &            - & - &- &-\\
			Deconfound$^{*}$\cite{tang2020long}           &    &  -               & -              &      -          &   -           & 51.8            & 62.7       &        48.8    &   31.6\\
			\midrule
			cRT\cite{kang2019decoupling}           &  \multirow{5}{*}{Learnable}  & 47.3             & 58.8           &    44.0         &  26.1         &  49.6        & 61.8          &        46.2    &   27.4\\
			cRT$^{*}$\cite{kang2019decoupling} &   & -                & -              &    -            &   -           &  49.7        & 60.4          &        46.8    &   29.3\\
			LWS\cite{kang2019decoupling}          &   & 47.7             & 57.1           &    45.2         &  29.3         &  49.9        & 60.2          &        47.2    &   30.3  \\
			\cmidrule{1-1}\cmidrule{3-10}
			\textbf{DisAlign}                     &   &  51.3 &  59.9  &  49.9   &  31.8      
			&  52.6  &  61.5 &  50.7  &  33.1\\
			\textbf{DisAlign$^{*}$}           &   &  \textbf{52.9}  &  61.3  &  52.2   &  31.4
			&  \textbf{53.4}  &   62.7  &  52.1   &  31.4\\
			
			\bottomrule
	\end{tabular}} 
		\caption{\textbf{Quantitative results on ImageNet-LT.} 
		$*$ denotes the model uses cosine classifier. 
		\textbf{R-50} and \textbf{X-50} means the ResNet-50 and ResNeXt-50, respectively.}
	\label{tab:imagenet_lt}
\end{table*}

 Given a train dataset $\mathcal{D}_{tr}=\{(\mathbf{x}_i,y_i)\}_{i=1}^N$, we introduce a calibration strategy based on distribution alignment between our model prediction $p_m(\cdot)$ and a reference distribution of classes that favors the balanced prediction.

Formally, denote the reference distribution as $p_{r}(y|\mathbf{x})$, we aim to minimize the expected KL-divergence between 
$p_{r}(y|\mathbf{x})$ and the model prediction $p_{m}(y|\mathbf{x})$ as follows:
\begin{align}
\mathcal{L} &= \mathbb{E}_{\mathcal{D}_{tr}}\left[\mathcal{KL}(p_r(y|\mathbf{x})||p_m(y|\mathbf{x}))\right]\label{eq:goal_1}\\
           &\approx -\frac{1}{N}\sum_{i=1}^N\left[ \sum_{y\in\mathcal{C}}p_r(y|\mathbf{x}_i)\log(p_m(y|\mathbf{x}_i))\right] +C\label{eq:goal_2}
\end{align}
where the expectation is approximated by an empirical average on $\mathcal{D}_{tr}$ and $C$ is a constant. 

In this work, we adopt a re-weighting approach~\cite{cui2019class} and introduce a generalized re-weight strategy for the alignment in order to exploit the class prior. Formally, we represent the reference distribution as a weighted empirical distribution on the training set,
\begin{align}
p_r(y=c|\mathbf{x}_i)= w_c\cdot\delta_c(y_i), \quad \forall c\in \mathcal{C}\label{eq:goal_3}
\end{align}
where $w_c$ is the class weight, and $\delta_c(y_i)$ is the Kronecker delta function(equals 1 if $y_i=c$, otherwise equals 0).
We then define the reference weight based on the empirical class frequencies $\mathbf{r}=[r_1,\cdots,r_K]$ on the training set:
\begin{align}
w_c=\frac{(1/r_c)^\rho}{\sum_{k=1}^K (1/r_k)^\rho}, \quad \forall c\in \mathcal{C}
\end{align}
where $\rho$ is a scale hyper-parameter to provide more flexibility in encoding class prior. Note that our scheme reduces to the instance-balance re-weight method with $\rho=0$, and to the class-balanced re-weight method with $\rho=1$. We illustrate the curve of re-weight coefficients based on ImageNet-LT dataset in Fig.~\ref{fig:exp_scale_illustration}. 

\subsection{Connection with Recent Work}\label{subsec:distribution_mismatch}
Below we discuss the connections between our proposed distribution alignment strategy and recent two-stage methods. Detailed comparison is reported in Tab.~\ref{tab:comparision_recent}.
Notably, Logit Adjustment\cite{menon2020long} and Deconfound\cite{tang2020long} introduce a hand-craft margin to adjust the distribution while keep the magnitude as 1.0, and incorporate the class prior directly in $r_i$ or $\mathbf{w}_i$ without re-training. LWS\cite{kang2019decoupling} and $\tau$-normalized\cite{kang2019decoupling} try to achieve a similar goal by learning a magnitude scale and discarding the margin adjustment. 

All these methods can be considered as the special cases of our DisAlign approach,
which provides a unified and simple form to model the distribution mismatch in a learnable way. 
Moreover, the resample based strategy is not easy to be applied for the instance-level (object detection/instance segmentation) or pixel-level (semantic segmentation) tasks, our generalized re-weight provides an alternative solution to incorporate the class prior in a simple and effective manner. Experimental results in Sec.~\ref{sec:exp} also demonstrate the strength of our method compared with the aforementioned works.

\begin{table}[t]
	\centering
	\resizebox{0.45\textwidth}{!}{
		\begin{tabular}{l|c|c|c|c}
			\toprule
			\multirow{2}{*}{\textbf{Method}}          & \multicolumn{4}{c}{\textbf{Align Method}}\\
			\cmidrule{2-5}
			& Type &  Balance  & Magnitude & Margin   \\
			\midrule
			Joint            &  -           &   -   & -          & -        \\\hline
			LWS\cite{kang2019decoupling}               & L   & CB-RS & $\alpha_j$        &             0                \\
			$\tau$-Normalized\cite{kang2019decoupling}  & H   & CB-RS & $1/||\textbf{w}_j||^\tau$   & 0         \\
			Logit Adjust\cite{menon2020long}  & H   & -           & 1.0        &   $-\lambda\log(r_j)$      \\
			Deconfound$^*$\cite{tang2020long}        & H & -   & 1.0        &   $-\lambda d(\mathbf{x},\mathbf{e})\mathbf{w}_j^\intercal \mathbf{e}$    \\
			\midrule
			\textbf{DisAlign}      & L   & G-RW           & $1+\sigma(\mathbf{x})\alpha_j$        &   $\sigma(\mathbf{x})\beta_j$    \\
			\textbf{DisAlign$^*$}    & L & G-RW             & $1+\sigma(\mathbf{x})\alpha_j$        &   $\sigma(\mathbf{x})\beta_j$    \\
			\toprule
	\end{tabular}}
		\caption{\textbf{Comparison with related methods.} $*$ denotes cosine classifier, \textbf{L}: learnable, \textbf{H}:hand-craft, \textbf{CB-RS}: class-balanced resampling, \textbf{G-RW}: generalized re-weight, $r_j$: class frequency for the $j$-th class, $\lambda$: hypper-parameter, $\mathbf{e}$: mean feature of training data, $d(\cdot)$: cosine distance.}
	\label{tab:comparision_recent}
\end{table}

\section{Experiments}\label{sec:exp}

In this section, we conduct a series of experiments to validate the effectiveness of our method. Below we present our experimental analysis and ablation study on the image classification task in Sec.~\ref{subsec:classification}, followed by our results on semantic segmentation task in Sec.~\ref{subsec:seg}.  In addition, we further evaluate our methods on object detection and instance segmentation tasks in Sec.~\ref{subsec:detection}. 

\subsection{Image Classification}\label{subsec:classification}
\paragraph{Experimental Details} To demonstrate our methods, we conduct experiments on three large-scale long-tail datasets, including ImageNet-LT~\cite{liu2019large}, iNaturalist 2018~\cite{van2018inaturalist}, and Places-LT~\cite{liu2019large}. 
We follow the experimental setting and implementation of \cite{kang2019decoupling}
\footnote{Detailed configuration and results are provided in the supplementary materials.}. 
For the ImageNet-LT dataset, we report performance with ResNet/ResNeXt-\{50,101,152\} as backbone, and mainly use ResNet-50 for ablation study. 
For iNaturalist 2018 and Places-LT, our comparisons are performed under the settings of ResNet-\{50,101,152\}.
\begin{table}[t!]
	\centering
	\resizebox{0.3\textwidth}{!}{
		\begin{tabular}{l|cc|cc}
			\toprule
				\multirow{2}{*}{\textbf{Method}} & \multicolumn{2}{c|}{\textbf{ResNet-50}}         & \multicolumn{2}{c}{\textbf{ResNet-152}} \\
				\cmidrule{2-5}
				                                 &  \textbf{90 E}      &  \textbf{200 E}    & \textbf{90 E}         & \textbf{200 E}\\
				\midrule
				LDAM\cite{cao2019learning} 								& 68.0         & -        & -         & -\\
				Baseline                            &  61.7            & 65.8  &  65.0     & 69.0\\
				Baseline$^*$                        &  64.8          & 66.2  &  67.3       & 69.0\\
				\midrule
				cRT\cite{kang2019decoupling}                              &  65.2        & 68.2  &  68.5         & 71.2\\
				$\tau$-norm\cite{kang2019decoupling}                      &  65.6          & 69.3 &  68.8        & 72.5\\
				LWS\cite{kang2019decoupling}                              &  65.9	     & 69.5 &  69.1         & 72.1\\
				BBN\cite{zhou2020bbn}            &  66.3	   & 69.6 & -               & - \\
				\midrule
				\textbf{DisAlign}                &  67.8    & \textbf{70.6} &  71.3          & \textbf{74.1}  \\ 
				\textbf{DisAlign$^*$}      &  \textbf{69.5}   & 70.2   &  \textbf{71.7}  & 72.8 \\
			\bottomrule
		\end{tabular}
	}
	\caption{\textbf{Average accuracy on iNaturalist-2018.} $*$ denotes the cosine classifier.}
	\label{tab:inaturalist}
\end{table}
\noindent\paragraph{Comparison with previous methods}
\textbf{1) ImageNet-LT.} We present the quantitative results for ImageNet-LT in Tab.~\ref{tab:imagenet_lt}. Our approach achieves \textbf{52.9\%} in per-class average accuracy based on ResNet-50 backbone and \textbf{53.4\%} based on ResNeXt-50, which outperform the state-of-the-art methods by a significant margin of \textbf{2.5\%} and \textbf{1.6\%}, respectively. 
\begin{table}[t!]
	\centering
	\resizebox{0.37\textwidth}{!}{
		\begin{tabular}{l|l|ccc}
			\toprule
				\multirow{2}{*}{\textbf{Method}} & \multicolumn{4}{c}{\textbf{ResNet-152}} \\
				\cmidrule{2-5}
				                                 &  \textbf{Average}                  &  \textbf{Many}                      & \textbf{Medium}                      &  \textbf{Few}\\
				\midrule
				Focal Loss\cite{liu2019large}                &  34.6                     & 41.1                       &    34.8                     &  22.4\\
				Range Loss\cite{liu2019large} &  35.1                     & 41.1                       &    35.4                     &  23.2\\
				OLTR\cite{liu2019large}       &  35.9                     & 44.7                       &    37.0                     &  25.3\\			
				Feature Aug\cite{chu2020feature} & 36.4 & 42.8& 37.5 &22.7\\		
				Baseline 						     &  30.2                     & \textbf{45.7}                       &    27.3                     &  8.2\\
				\midrule
				NCM & 36.4                     & 40.4       & 37.1 & 27.3\\
				cRT & 36.7                     & 42.0       & 37.6 & 24.9\\
				LWS & 37.6                     & 40.6       & 39.1 & 28.6\\
				$\tau$-norm          &  37.9   & 37.8       & 40.7 & \textbf{31.8}\\
				\midrule
				\textbf{DisAlign}                & \textbf{39.3} & 40.4 & \textbf{42.4} & 30.1 \\
			\bottomrule
		\end{tabular}
	}
	\caption{\textbf{Results on Place365-LT with ResNet-152.}}
	\label{tab:places}
\end{table}
\noindent\textbf{2) iNaturalist.} 
In Tab.~\ref{tab:inaturalist}, our method DisAlign with cosine classifier achieves \textbf{69.5\%} per-class average accuracy using ResNet-50 backbone and 90 epochs of training, surpassing the prior art LDAM by a large margin at \textbf{1.5\%}. It also shows that our performance can be further improved with larger backbone and/or more training epochs.
\noindent\textbf{3) Places-LT.} In Tab.~\ref{tab:places}, we show the experimental results under the same setting as \cite{kang2019decoupling} on Places-LT. Our method achieves \textbf{39.3\%} per-class average accuracy based on ResNet-152, with a notable performance gain at \textbf{1.4\%} over the prior methods. We also report the detailed performance of these three datasets with ResNet-\{50,101,152\} in the supplementary materials. 

\begin{table}[t!]
	\centering
	\resizebox{0.38\textwidth}{!}{
		\begin{tabular}{c|c|c|c|ccc}
			\toprule
				\textbf{GR}  & \textbf{MT}         & \textbf{MG}                &  \textbf{Average}                   & \textbf{Many}              & \textbf{Medium}         & \textbf{Few} \\
				\midrule
				\xmark & \xmark                           & \xmark &   41.6	           & 64.0           &    33.8         &  5.8\\
				\midrule
				\cmark & \cmark                           & \xmark &   50.1            & 60.4 &    48.0         & 28.8\\
				\cmark & \xmark                           & \cmark &  49.9            & 63.9 &    46.9         & 21.2\\
				\cmark & \cmark                           & \cmark & 51.3 & 59.9& 49.9 &31.8\\
			\bottomrule
		\end{tabular}
	}
		\caption{\textbf{Ablation study of DisAlign.} \textbf{GR} means the generalized reweight strategy.  \textbf{MT} means the learnable magnitude parameter (1+$\sigma(\mathbf{x})\alpha$) and \textbf{MG} is the learnable margin parameter  $\sigma(\mathbf{x})\beta$.}
	\label{tab:model_ablation}
\end{table}

\begin{table*}[h!]
\centering
\resizebox{0.8\textwidth}{!}{
\begin{tabular}{l|l|c||l|lll||l|lll}
\toprule
\multirow{2}{*}{\textbf{Framework}} &\multirow{2}{*}{\textbf{Method}} & \multirow{2}{*}{\textbf{B}} & \multicolumn{4}{c||}{\textbf{Mean IoU(\%)}} & \multicolumn{4}{c}{\textbf{Mean Accuracy(\%)}}\\
\cmidrule{4-11}
       &       & & \textbf{Average} & \textbf{Head} & \textbf{Body} & \textbf{Tail} &  \textbf{Average} & \textbf{Head} & \textbf{Body} & \textbf{Tail} \\
\midrule
\multirow{6}{*}{FCN\cite{shelhamer2017fully}} & Baseline & \multirow{2}{*}{R-50}    & 38.1  &  64.6   &  40.0   & 29.6 
															  & 46.3  &  78.6   &  49.3   & 35.4 \\
& \textbf{DisAlign}         &                       & 40.1\more{(+2.0)}  & 65.0\more{(+0.4)}  & 42.8\more{(+2.8)} & 31.3\more{(+1.7)}   
															  & 51.4\more{(+5.1)}  & 78.6\small{(+0.0)} & 56.1\more{(+6.8)} & 40.6\more{(+5.2)}\\
\cmidrule{2-11}
& Baseline &  \multirow{2}{*}{R-101}     &  41.4  &   67.0  &  43.3   & 33.2 
																   &  50.2  &   80.6  &  52.9   & 40.1 \\
&\textbf{DisAlign}         &                             &  43.7\more{(+2.3)}  & 67.4\more{(+0.4)}  & 46.1\more{(+2.8)} & 35.7\more{(+2.5)}   
														           &  55.9\more{(+5.7)}  & 80.6\small{(+0.0)} & 59.7\more{(+6.8)} & 46.4\more{(+6.3)}  \\
\cmidrule{2-11}
&Baseline  &  \multirow{2}{*}{S-101}          & 46.2  & 67.6   & 48.0  & 39.1 
																	  & 57.3  & 79.4   & 61.7  & 48.2   \\
&\textbf{DisAlign}         &                              & 46.9\more{(+0.7)}  & 67.7\more{(+0.1)}  & 48.2\more{(+0.2)}  & 40.3\more{(+1.2)}
																      & 60.1\more{(+2.8)} &  79.7\more{(+0.3)}  & 64.2\more{(+2.5)}  & 51.9\more{(+3.7)} \\
\midrule
\multirow{6}{*}{DeepLabV3$^{+}$\cite{deeplabv3plus2018}} & Baseline & \multirow{2}{*}{R-50}   &  44.9  &  67.7  &   48.3  &   36.4 
																		  &  55.0  &  80.1  &   60.8  &   44.1 \\
&\textbf{DisAlign}        &                              & 45.7\more{(+0.8)} & 67.7\small{(+0.0)} & 48.6\more{(+0.3)} & 37.8\more{(+1.4)}
																          &  57.3\more{(+2.3)} & 80.8\more{(+0.7)} & 63.0\more{(+2.2)} & 46.9\more{(+2.8)} \\
\cmidrule{2-11}
& Baseline &\multirow{2}{*}{R-101}  &  46.4 & 68.7  & 49.0 & 38.4 & 56.7 & 80.9 & 61.5 & 46.7 \\
&\textbf{DisAlign}        &                        &  47.1\more{(+0.7)} & 68.7\small{(+0.0)} & 49.4\more{(+0.4)} & 39.6\more{(+1.2)}
																          &  59.5\more{(+2.8)} & 81.4\more{(+0.5)} & 64.2\more{(+2.7)} & 50.3\more{(+3.6)} \\
\cmidrule{2-11}
&  Baseline &\multirow{2}{*}{S-101}  &  47.3 & 69.0 & 49.7 & 39.7
																	      &  58.1 & 80.8 & 63.4 & 48.2\\
&\textbf{DisAlign}        &                       &  47.8\more{(+0.5)} & 68.9\small{(-0.1)} & 49.8\more{(+0.1)} & 40.7\more{(+1.0)} 
																		  &  60.1\more{(+2.0)} & 81.0\more{(+0.2)}  & 65.5\more{(+2.1)} & 52.0\more{(+3.8)}\\
\bottomrule
\end{tabular}}
\caption{\textbf{Performance of semantic segmentation  on ADE-20K:} All baseline models are trained with an image size of 512$\times$512 and 160K iteration in total. \textbf{B} is backbone network(R-50, R-101, S-101 denote ResNet-50, ResNet-101 and ResNeSt-101, respectively).}
\label{tab:segmentation}
\end{table*}

\begin{figure}
    \centering
    \includegraphics[width=0.8\linewidth]{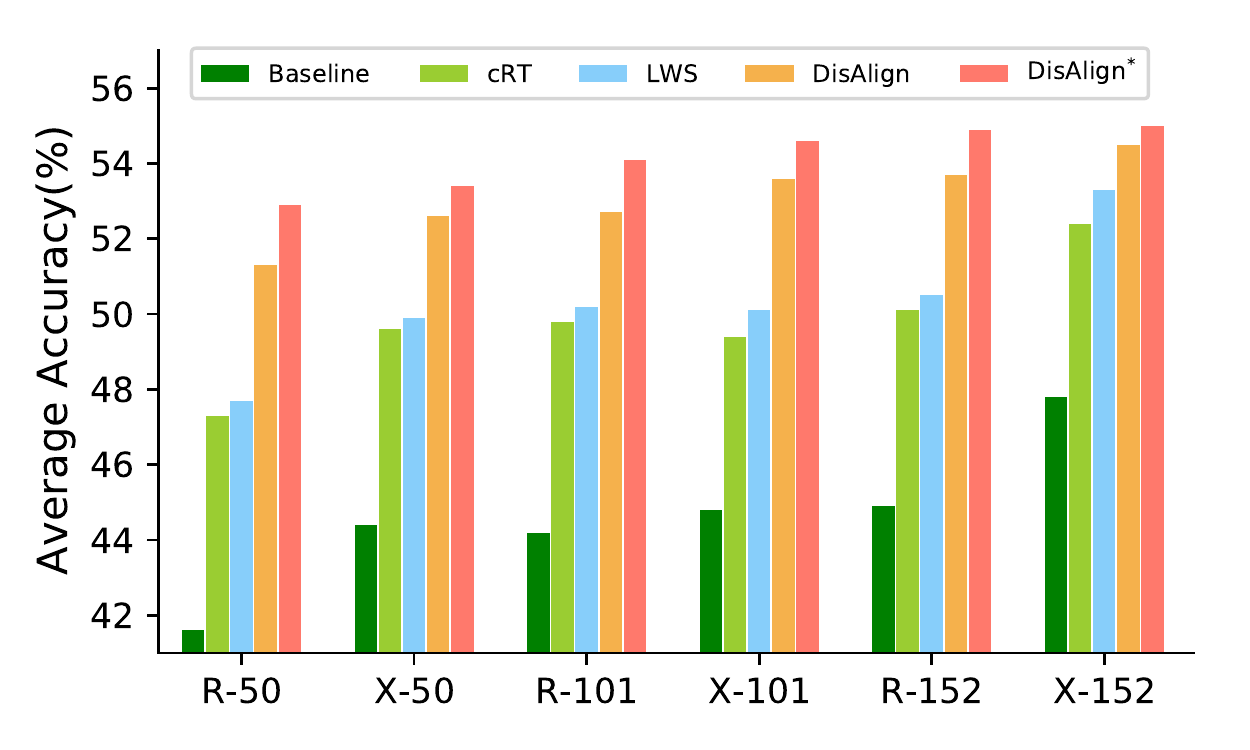}
    \caption{\textbf{Performance of DisAlign with different backbone on ImageNet-LT.} Detailed results will be reported in supplementary material.}
    \label{fig:backbones}
\end{figure}

\paragraph{Ablation Study}
\textbf{1) Different Backbone:} We validate our method on different types of backbone networks, ranging from ResNet-\{50,101,152\} to ResNeXt-\{50, 101, 152\}, reported in Fig.~\ref{fig:backbones}. 
Our method achieves \textbf{54.9\%} with ResNet-152, and \textbf{55.0\%} with ResNeXt-152. It's worth noting that even when adopting stronger backbones, the gain of DisAlign compared to the state-of-the-art methods is still significant. This demonstrates that our DisAlign is complementary to the capacity of  backbone networks.
\begin{figure}
    \centering

    \includegraphics[width=0.8\linewidth]{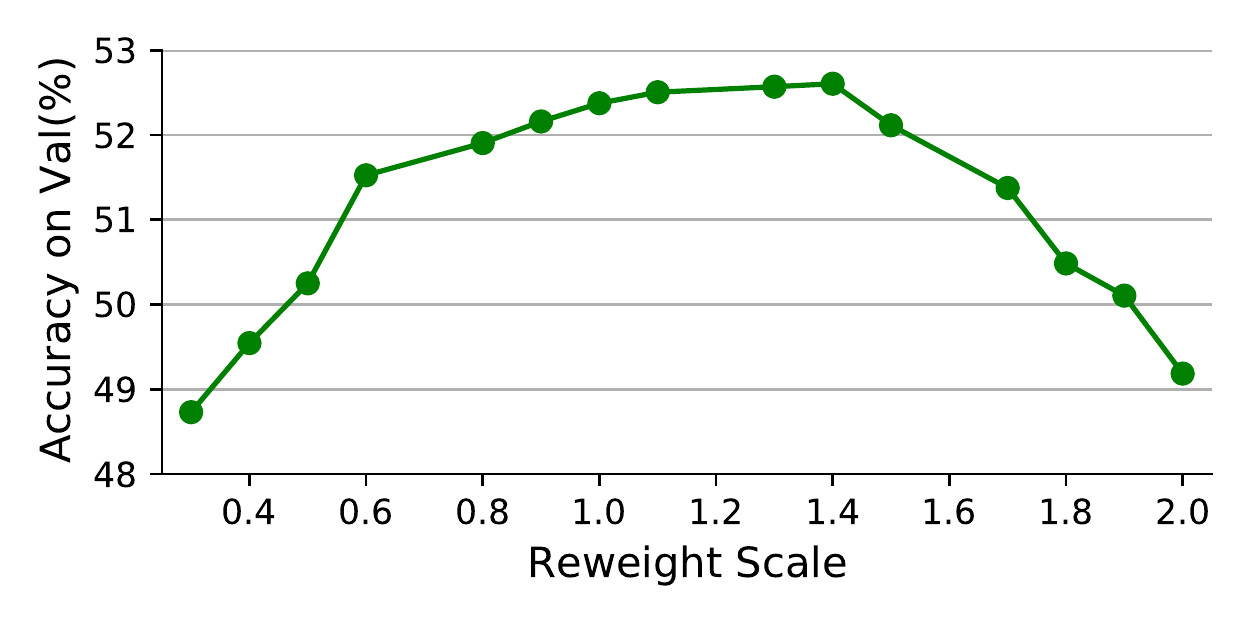}
    \caption{\textbf{Effects of the different generalized re-weight scale.} Performance is reported on ImageNet-LT val split.}
    \label{fig:exp_scale_ablation}
\end{figure}
\noindent
\textbf{2) Model Components:}
We conduct a series of ablation studies to evaluate the importance of each component used in our DisAlign method. Tab.~\ref{tab:model_ablation} summarizes the results of our ablation experiments, in which we compare our full model with several partial model settings. 
From the table, we find the learnable magnitude has a significant improvement compared with baseline and the learnable margin also achieves competitive results at 49.9\%, which demonstrate the effectiveness of individual modules in our design.
\noindent\textbf{3) Generalized Re-weight Scale}
We also investigate the influence of the generalized re-weight scale on the validation set of ImageNet-LT and plot the accuracy-scale curve in Fig.~\ref{fig:exp_scale_ablation}. It is evident that adjusting generalized reweight is able to achieve significant performance improvement. Moreover, we find the setting of $\rho>1$ is able to outperform the class-balanced re-weight ($\rho=1$), which indicates that the generalized re-weight is more effective in coping with long-tail distributions.


\begin{table}[t!]
	\centering
	\resizebox{0.43\textwidth}{!}{
		\begin{tabular}{c|l|ll|ll}
			\toprule
			\multirow{2}{*}{\textbf{B}} & \multirow{2}{*}{\textbf{Method}}     & \multicolumn{2}{c|}{\textbf{Mask R-CNN}} &\multicolumn{2}{c}{\textbf{Cascade R-CNN}}  \\
			\cmidrule{3-6}
			&  &$\text{\textbf{AP}}_{bbox}$ & $\text{\textbf{AP}}_{mask}$   & $\text{\textbf{AP}}_{bbox}$ & $\text{\textbf{AP}}_{mask}$   \\
			\midrule
			\multirow{4}{*}{R-50 } & Baseline        & 20.8                 & 21.2               &  25.2            & 23.0         \\
			& \textbf{DisAlign}                   & 23.9\more{(+3.1)}   & 24.2\more{(+3.0)}   & 28.7\more{(+3.5)} & 26.1\more{(+3.1)}      \\
			\cmidrule{2-6}
			& Baseline$^{*}$                          & 22.8                & 23.8                &  28.8              & 26.2         \\
			& \textbf{DisAlign$^{*}$}             & 25.6\more{(+2.8)}   & 26.3\more{(+2.5)}   &  32.2\more{(+3.4)} & 29.4\more{(+3.2)} \\
			\midrule
			\multirow{4}{*}{R-101 } & Baseline       & 22.2                & 22.6                &  26.1              & 24.0         \\
			& \textbf{DisAlign}                   & 25.6\more{(+3.4)}   & 25.8\more{(+3.2)}   &  29.7\more{(+3.6)} & 27.3\more{(+3.3)}\\
			\cmidrule{2-6}
			& Baseline$^{*}$                          & 24.5                & 25.1                &  30.4              & 28.1        \\
			& \textbf{DisAlign$^{*}$}             & 27.5\more{(+3.0)}   & 28.2\more{(+3.1)}       &  33.7\more{(+3.3)} & 30.9\more{(+2.8)}\\
			\midrule
			\multirow{4}{*}{X-101 } & Baseline     & 24.5                & 25.0               &  28.4              & 26.1         \\
			& \textbf{DisAlign}                  & 26.8\more{(+2.3)}   & 27.4\more{(+2.4)}   &  31.3\more{(+2.9)} & 28.7\more{(+2.6)} \\
			\cmidrule{2-6}
			& Baseline$^{*}$                      & 26.9                & 27.7                &  32.6              & 29.8\\
			& \textbf{DisAlign$^{*}$}             & 29.5\more{(+2.6)}   & 30.0\more{(+2.3)}   &  34.7\more{(+2.1)} & 31.8\more{(+2.0)} \\
			\bottomrule
	\end{tabular}}
	\caption{\textbf{Results on LVIS v0.5 dataset with different backbones and different architectures.} The results are reported based on the Detectron2\cite{wu2019detectron2,zhu2020cvpods} framework. We refer the reader to the supplementary material for the detailed comparison with the state of art.}
	\label{tab:lvis_ablation}
\end{table}

\subsection{Semantic Semgnetaion on ADE20k Dataset}\label{subsec:seg}
To further validate our method, we apply DisAlign strategy to segmentation networks and report our performance on the semantic segmentation benchmark, ADE20k~\cite{zhou2017scene}. 

\begin{table*}
\centering
    \resizebox{0.75\textwidth}{!}{
\begin{tabular}{l|l|c|ccc|c|ccc}
\toprule
  \multirow{2}{*}{\textbf{Pre-Train}}& \multirow{2}{*}{\textbf{Method}} & \multicolumn{4}{c|}{\textbf{BBox AP}} & \multicolumn{4}{c}{\textbf{Mask AP}} \\
  \cmidrule{3-10}
  &  & $\text{\textbf{AP}}_{bbox}$ & $\text{\textbf{AP}}_{bbox}^r$ & $\text{\textbf{AP}}_{bbox}^c$ & $\text{\textbf{AP}}_{bbox}^f$ & $\text{\textbf{AP}}_{mask}$& $\text{\textbf{AP}}_{mask}^r$ & $\text{\textbf{AP}}_{mask}^c$ & $\text{\textbf{AP}}_{mask}^f$ \\
\midrule
 \multirow{8}{*}{ImageNet}	 & Baseline 	                & 20.8  &  3.3   &  19.5  & 29.4     & 21.2 &   3.7  & 21.6 & 28.4  \\
          & Baseline$^{*}$ 	            &                    22.8  &  10.2  &  21.1  & 30.1     & 23.8 &   11.5 & 23.7 & 28.9 \\
          &Focal Loss\cite{lin2017focal}                                      & 21.9  & -      & -      & -        & 21.0 & 9.3 & 21.0 & 25.8 \\
          &SimCal\cite{wang2020devil}				                & 22.6  &  13.7  &  20.6  & 28.7     & 23.4 &   16.4  & 22.5 & 27.2 \\ 
          &LST\cite{hu2020learning} & 22.6  & - & - &- & 23.0 & - & - & -\\
          &RFS\cite{gupta2019lvis}				                  & 23.6  &	12.8  &  22.3  & 29.4     & 24.3 &   14.6 & 24.0 & 28.5\\
          &EQL\cite{tan2020equalization}			              & 23.3  &  -     &  -     & -        & 22.8 &   11.3 & 24.7 & 25.1 \\
\cmidrule{2-10}
         &  \textbf{DisAlign}                                 & 23.9   & 7.5    & 25.0   & 29.1     & 24.3 & 8.5 & 26.3 & 28.1\\
          & \textbf{DisAlign$^*$}                          & \textbf{25.6} &   13.7  &  25.6  &  30.5 & \textbf{26.3} & 14.9 &  27.6  &   29.2  \\
\midrule
 \multirow{6}{*}{COCO} 	&  Baseline			  &  22.8 &  2.6  &  21.8 & 32.0     & 23.9 & 2.8 & 23.4 & 30.5  \\
   & Baseline$^*$ 	  &                        25.0 &  10.2  & 23.9 & 32.3     & 25.3 & 11.0 & 25.5 & 30.7 \\
  &  GroupSoftmax\cite{li2020overcoming}       & 25.8 & 15.0 & 25.5 & 30.4       & 26.3 &  18.0 & 26.9 & 28.7 \\
\cmidrule{2-10}
  & \textbf{DisAlign}            &    25.5 & 8.2 & 26.3 & 32.4 & 25.7 & 9.4 & 27.6 & 29.7 \\
  & \textbf{DisAlign$^*$}          &	   \textbf{27.6} & 14.8   &  27.9  &  32.4    & \textbf{27.9} & 16.2 & 29.3  &  30.8 \\
\bottomrule
\end{tabular}}
\caption{\textbf{Comparison with the-state-of-art on LVIS with Mask-R-CNN-FPN(ResNet-50 backbone).} All results are evaluated on the LVIS v0.5 validation set with the score threshold at 0.0001. ($*$ denotes cosine classifier for bbox classification.)}
\label{tab:lvis_sota}
\end{table*}

\begin{table*}
	\centering
	\resizebox{0.78\textwidth}{!}{
	  \begin{tabular}{c|l|c|ccc|c|ccc}
		\toprule
	  
		  \multirow{2}{*}{\textbf{Backbone}} & \multirow{2}{*}{\textbf{Method}}  &   \multicolumn{4}{c|}{\textbf{BBox AP}} & \multicolumn{4}{c}{\textbf{Mask AP}} \\
		 \cmidrule{3-10}
		  &  &$\text{\textbf{AP}}_{bbox}$ & $\text{\textbf{AP}}_{bbox}^r$ & $\text{\textbf{AP}}_{bbox}^c$ & $\text{\textbf{AP}}_{bbox}^f$ & $\text{\textbf{AP}}_{mask}$& $\text{\textbf{AP}}_{mask}^r$ & $\text{\textbf{AP}}_{mask}^c$ & $\text{\textbf{AP}}_{mask}^f$ \\
	\midrule
	\multirow{2}{*}{ResNet-50 } & Baseline$^{*}$                          & 26.5 &	8.7 & 25.0 & 36.0 &	23.5 &	8.1 &	22.4 &	31.5   \\
	& \textbf{DisAlign$^{*}$}                 & 30.5 &  17.9 & 30.1 & 36.5 & 27.0 & 15.7 & 27.0 & 31.9 \\
	\midrule
	\multirow{4}{*}{ResNet-101 } & De-confound\cite{tang2020long} & 25.8 & - & - & - & 23.5 & 5.2 & 22.7 & 32.3\\
	& De-confound TDE\cite{tang2020long}   & 30.0 & - & - & - & 27.1 & 16.0 & 26.9 & 32.1 \\
		 \cmidrule{2-10}
	& Baseline$^{*}$                         & 28.9 & 11.8 & 27.7  & 37.8 &	25.6 &	10.5 &	24.9 &	33.0 \\
	& \textbf{DisAlign$^{*}$}                & 32.7 & 20.5 & 32.8  & 38.1 & 28.9 &  18.0 & 29.3 & 33.3  \\
	\midrule
	\multirow{2}{*}{ResNeXt-101 } & Baseline$^{*}$                      & 30.7 & 14.2 & 29.3 & 39.6 & 27.3 & 13.0 & 26.4 & 34.6\\
	& \textbf{DisAlign$^{*}$}             & 33.7 & 21.4 & 33.1 & 39.7 & 29.7 & 18.4 & 29.7 & 34.7 \\
	\bottomrule
	\end{tabular}}
	\caption{\textbf{Results on LVIS v1.0 dataset with Cascade R-CNN.} * denotes cosine classifier head.}
	\label{tab:lvis_v1}
	\end{table*}

\noindent\paragraph{Dataset and Implementation Details}
Follow a similar protocol as in image classification, we divide the 150 categories into 3 subsets according to the percentage of pixels in every category over the entire dataset. Specifically, we define three disjoint subsets as follows: \textit{head classes} (each with more than 1.0\% of total pixels), \textit{body classes} (each with a percentage ranging from 0.1\% to 1\% of total pixels) and \textit{tail classes} (each with less than 0.1\% of total pixels). \footnote{The complete list of the split is reported in supplementary material.}


\noindent\paragraph{Quantitative Results} We evaluate our method using two widely-adopted segmentation models (FCN~\cite{shelhamer2017fully} and DeepLabV3$^{+}$~\cite{deeplabv3plus2018}) based on different backbone networks, ranging from ResNet-50, ResNet-101 to the latest ResNeSt-101, and report the performance in Tab.~\ref{tab:segmentation}. 
Our method achieves \textbf{2.0} and \textbf{2.3} improvement in mIoU using FCN-8s with ResNet-50 and ResNet-101, respectively. The performance on the body and tail are improved significantly. Moreover, our method outperforms the baseline with large margin at \textbf{5.7} in mean accuracy with ResNet-101 backbone. Even with a stronger backbone: \textbf{ResNeSt-101}~\cite{zhang2020resnest}, our method also achieves \textbf{0.7} mIoU and \textbf{2.8} improvement in mean accuracy, where the tail categories have a performance gain of \textbf{1.2} in mIoU and \textbf{3.7} in mean accuracy.
We further validate our method using DeepLabV3$^{+}$, which is a more powerful semantic segmentation model. Our DisAlign improves the performance of DeepLabV3$^{+}$ by a margin of \textbf{0.5} based on ResNeSt-101, which achieves the new state-of-the-art (\textbf{47.8} in mIoU) on the ADE20k dataset.

\subsection{Object Detection and Instance Segmentation}\label{subsec:detection}

\noindent\paragraph{Experimental Configuration}

We conduct experiments on LVIS~\cite{gupta2019lvis} dataset. 
For evaluation, we use a COCO-style average precision (AP) metric that averages over categories and different box/mask IoU threshold~\cite{lin2014microsoft}.

\noindent\paragraph{Quantitative Results and Ablation Study}
We first compare our method with recent work and report quantitative results in Tab.~\ref{tab:lvis_sota}. We find our DisAlign with cosine classifier head achieves \textbf{25.6} in $\text{AP}_{bbox}$, and \textbf{26.3} in $\text{AP}_{mask}$ when applied to the Mask R-CNN+FPN with the ImageNet pre-trained ResNet-50 backbone. Moreover, our strategy can be further improved to achieve \textbf{27.6} in $\text{AP}_{bbox}$ and \textbf{27.9} in $\text{AP}_{mask}$ based on the COCO pre-trained model. In both cases, our method is able to maintain the performance of the frequent (also called \textit{head}) categories, and gain significant improvement on common (also called \textit{body}) and rare (also called \textit{tail}) categories.
We also report performance with more power detection framework (\textit{e.g.}Cascade R-CNN) and stronger backbones (\textit{e.g.} ResNet-50/101, and ResNeXt-101) in Tab.~\ref{tab:lvis_ablation} and Tab.~\ref{tab:lvis_v1}. It is worth noting that even with the stronger backbones or frameworks, the performance gain of our DisAlign over the baseline is still significant.
\vspace{-0.2em}
\section{Conclusion}  
In this paper, we have presented a unified two-stage learning strategy for the large-scale long-tail visual recognition tasks. To tackle the biased label prediction, we develop a confidence-aware distribution alignment method to calibrate initial classification predictions. In particular, we design a generalized re-weight scheme to leverage the category prior for the alignment process. Extensive experiments show that our method outperforms previous works with a large margin on a variety of visual recognition tasks(image classification, semantic segmentation, and object detection/segmentation).

{\small
\bibliographystyle{ieee_fullname}
\bibliography{egbib}

\begin{thebibliography}{10}\itemsep=-1pt

\bibitem{bengio2015sharing}
Samy Bengio.
\newblock Sharing representations for long tail computer vision problems.
\newblock In {\em Proceedings of the ACM on International Conference on
  Multimodal Interaction}, 2015.

\bibitem{buda2018systematic}
Mateusz Buda, Atsuto Maki, and Maciej~A Mazurowski.
\newblock A systematic study of the class imbalance problem in convolutional
  neural networks.
\newblock {\em Neural Networks}, 2018.

\bibitem{cao2019learning}
Kaidi Cao, Colin Wei, Adrien Gaidon, Nikos Arechiga, and Tengyu Ma.
\newblock Learning imbalanced datasets with label-distribution-aware margin
  loss.
\newblock In {\em Advances in Neural Information Processing Systems(NeurIPS)},
  2019.

\bibitem{chawla2002smote}
Nitesh~V Chawla, Kevin~W Bowyer, Lawrence~O Hall, and W~Philip Kegelmeyer.
\newblock Smote: synthetic minority over-sampling technique.
\newblock {\em Journal of Artificial Intelligence Research}, 2002.

\bibitem{chen2018deeplab}
Liang-Chieh Chen, George Papandreou, Iasonas Kokkinos, Kevin Murphy, and Alan~L
  Yuille.
\newblock Deeplab: Semantic image segmentation with deep convolutional nets,
  atrous convolution, and fully connected crfs.
\newblock {\em IEEE Transactions on Pattern Analysis and Machine
  Intelligence(TPAMI)}, 2018.

\bibitem{chen2017rethinking}
Liang-Chieh Chen, George Papandreou, Florian Schroff, and Hartwig Adam.
\newblock Rethinking atrous convolution for semantic image segmentation.
\newblock {\em arXiv preprint}, 2017.

\bibitem{deeplabv3plus2018}
Liang-Chieh Chen, Yukun Zhu, George Papandreou, Florian Schroff, and Hartwig
  Adam.
\newblock Encoder-decoder with atrous separable convolution for semantic image
  segmentation.
\newblock In {\em Proceedings of the European Conference on Computer
  Vision(ECCV)}, 2018.

\bibitem{chu2020feature}
Peng Chu, Xiao Bian, Shaopeng Liu, and Haibin Ling.
\newblock Feature space augmentation for long-tailed data.
\newblock In {\em Proceedings of the European Conference on Computer
  Vision(ECCV)}, 2020.

\bibitem{cui2019class}
Yin Cui, Menglin Jia, Tsung-Yi Lin, Yang Song, and Serge Belongie.
\newblock Class-balanced loss based on effective number of samples.
\newblock In {\em Proceedings of the IEEE/CVF Conference on Computer Vision and
  Pattern Recognition(CVPR)}, 2019.

\bibitem{deng2009imagenet}
Jia Deng, Wei Dong, Richard Socher, Li-Jia Li, Kai Li, and Li Fei-Fei.
\newblock Imagenet: A large-scale hierarchical image database.
\newblock In {\em Proceedings of the IEEE/CVF Conference on Computer Vision and
  Pattern Recognition(CVPR)}, 2009.

\bibitem{drummond2003c4}
Chris Drummond, Robert~C Holte, et~al.
\newblock C4. 5, class imbalance, and cost sensitivity: why under-sampling
  beats over-sampling.
\newblock In {\em Workshop on Learning from Imbalanced Datasets, II}, 2003.

\bibitem{gao2018solution}
Yuan Gao, Xingyuan Bu, Yang Hu, Hui Shen, Ti Bai, Xubin Li, and Shilei Wen.
\newblock Solution for large-scale hierarchical object detection datasets with
  incomplete annotation and data imbalance.
\newblock {\em arXiv preprint}, 2018.

\bibitem{girshick2015fast}
Ross Girshick.
\newblock Fast r-cnn.
\newblock In {\em Proceedings of the IEEE International Conference on Computer
  Vision(ICCV)}, 2015.

\bibitem{girshick2014rich}
Ross Girshick, Jeff Donahue, Trevor Darrell, and Jitendra Malik.
\newblock Rich feature hierarchies for accurate object detection and semantic
  segmentation.
\newblock In {\em Proceedings of the IEEE/CVF Conference on Computer Vision and
  Pattern Recognition(CVPR)}, 2014.

\bibitem{gupta2019lvis}
Agrim Gupta, Piotr Dollar, and Ross Girshick.
\newblock Lvis: A dataset for large vocabulary instance segmentation.
\newblock In {\em Proceedings of the IEEE/CVF Conference on Computer Vision and
  Pattern Recognition(CVPR)}, 2019.

\bibitem{han2005borderline}
Hui Han, Wen-Yuan Wang, and Bing-Huan Mao.
\newblock Borderline-smote: a new over-sampling method in imbalanced data sets
  learning.
\newblock In {\em International Conference on Intelligent Computing}, 2005.

\bibitem{he2017mask}
Kaiming He, Georgia Gkioxari, Piotr Doll{\'a}r, and Ross Girshick.
\newblock Mask r-cnn.
\newblock In {\em Proceedings of the IEEE International Conference on Computer
  Vision(ICCV)}, 2017.

\bibitem{hu2020learning}
Xinting Hu, Yi Jiang, Kaihua Tang, Jingyuan Chen, Chunyan Miao, and Hanwang
  Zhang.
\newblock Learning to segment the tail.
\newblock In {\em Proceedings of the IEEE/CVF Conference on Computer Vision and
  Pattern Recognition(CVPR)}, 2020.

\bibitem{huang2016learning}
Chen Huang, Yining Li, Chen Change~Loy, and Xiaoou Tang.
\newblock Learning deep representation for imbalanced classification.
\newblock In {\em Proceedings of the IEEE/CVF Conference on Computer Vision and
  Pattern Recognition(CVPR)}, 2016.

\bibitem{kang2019decoupling}
Bingyi Kang, Saining Xie, Marcus Rohrbach, Zhicheng Yan, Albert Gordo, Jiashi
  Feng, and Yannis Kalantidis.
\newblock Decoupling representation and classifier for long-tailed recognition.
\newblock {\em International Conference on Learning Representations(ICLR)},
  2020.

\bibitem{khan2019striking}
Salman Khan, Munawar Hayat, Syed~Waqas Zamir, Jianbing Shen, and Ling Shao.
\newblock Striking the right balance with uncertainty.
\newblock In {\em Proceedings of the IEEE/CVF Conference on Computer Vision and
  Pattern Recognition(CVPR)}, 2019.

\bibitem{khan2017cost}
Salman~H Khan, Munawar Hayat, Mohammed Bennamoun, Ferdous~A Sohel, and Roberto
  Togneri.
\newblock Cost-sensitive learning of deep feature representations from
  imbalanced data.
\newblock {\em IEEE Transactions on Neural Networks and Learning
  Systems(TNNLS)}, 2017.

\bibitem{krizhevsky2012imagenet}
Alex Krizhevsky, Ilya Sutskever, and Geoffrey~E Hinton.
\newblock Imagenet classification with deep convolutional neural networks.
\newblock In {\em Advances in Neural Information Processing Systems(NeurIPS)},
  2012.

\bibitem{li2020overcoming}
Yu Li, Tao Wang, Bingyi Kang, Sheng Tang, Chunfeng Wang, Jintao Li, and Jiashi
  Feng.
\newblock Overcoming classifier imbalance for long-tail object detection with
  balanced group softmax.
\newblock In {\em Proceedings of the IEEE/CVF Conference on Computer Vision and
  Pattern Recognition(CVPR)}, 2020.

\bibitem{lin2017feature}
Tsung-Yi Lin, Piotr Doll{\'a}r, Ross Girshick, Kaiming He, Bharath Hariharan,
  and Serge Belongie.
\newblock Feature pyramid networks for object detection.
\newblock In {\em Proceedings of the IEEE/CVF Conference on Computer Vision and
  Pattern Recognition(CVPR)}, 2017.

\bibitem{lin2017focal}
Tsung-Yi Lin, Priya Goyal, Ross Girshick, Kaiming He, and Piotr Doll{\'a}r.
\newblock Focal loss for dense object detection.
\newblock In {\em Proceedings of the IEEE International Conference on Computer
  Vision(ICCV)}, 2017.

\bibitem{lin2014microsoft}
Tsung-Yi Lin, Michael Maire, Serge Belongie, James Hays, Pietro Perona, Deva
  Ramanan, Piotr Doll{\'a}r, and C~Lawrence Zitnick.
\newblock Microsoft coco: Common objects in context.
\newblock In {\em Proceedings of the European Conference on Computer
  Vision(ECCV)}, 2014.

\bibitem{liu2019large}
Ziwei Liu, Zhongqi Miao, Xiaohang Zhan, Jiayun Wang, Boqing Gong, and Stella~X
  Yu.
\newblock Large-scale long-tailed recognition in an open world.
\newblock In {\em Proceedings of the IEEE/CVF Conference on Computer Vision and
  Pattern Recognition(CVPR)}, 2019.

\bibitem{mahajan2018exploring}
Dhruv Mahajan, Ross Girshick, Vignesh Ramanathan, Kaiming He, Manohar Paluri,
  Yixuan Li, Ashwin Bharambe, and Laurens van~der Maaten.
\newblock Exploring the limits of weakly supervised pretraining.
\newblock In {\em Proceedings of the European Conference on Computer
  Vision(ECCV)}, 2018.

\bibitem{menon2020long}
Aditya~Krishna Menon, Sadeep Jayasumana, Ankit~Singh Rawat, Himanshu Jain,
  Andreas Veit, and Sanjiv Kumar.
\newblock Long-tail learning via logit adjustment.
\newblock {\em arXiv preprint}, 2020.

\bibitem{mmsegmentation}
Open MMLab.
\newblock Mmsegmentation.
\newblock \url{https://github.com/open-mmlab/mmsegmentation}, 2020.

\bibitem{paszke2019pytorch}
Adam Paszke, Sam Gross, Francisco Massa, Adam Lerer, James Bradbury, Gregory
  Chanan, Trevor Killeen, Zeming Lin, Natalia Gimelshein, Luca Antiga, et~al.
\newblock Pytorch: An imperative style, high-performance deep learning library.
\newblock In {\em Advances in Neural Information Processing Systems(NeurIPS)},
  2019.

\bibitem{qi2018low}
Hang Qi, Matthew Brown, and David~G Lowe.
\newblock Low-shot learning with imprinted weights.
\newblock In {\em Proceedings of the IEEE/CVF Conference on Computer Vision and
  Pattern Recognition(CVPR)}, 2018.

\bibitem{ren2018learning}
Mengye Ren, Wenyuan Zeng, Bin Yang, and Raquel Urtasun.
\newblock Learning to reweight examples for robust deep learning.
\newblock In {\em International Conference on Machine Learning(ICML)}, 2018.

\bibitem{ren2015faster}
Shaoqing Ren, Kaiming He, Ross Girshick, and Jian Sun.
\newblock Faster r-cnn: Towards real-time object detection with region proposal
  networks.
\newblock In {\em Advances in Neural Information Processing Systems(NeurIPS)},
  2015.

\bibitem{shelhamer2017fully}
Evan Shelhamer, Jonathan Long, and Trevor Darrell.
\newblock Fully convolutional networks for semantic segmentation.
\newblock {\em IEEE Transactions on Pattern Analysis and Machine
  Intelligence(TPAMI)}, 2017.

\bibitem{shen2016relay}
Li Shen, Zhouchen Lin, and Qingming Huang.
\newblock Relay backpropagation for effective learning of deep convolutional
  neural networks.
\newblock In {\em Proceedings of the European Conference on Computer
  Vision(ECCV)}, 2016.

\bibitem{tan2020equalization}
Jingru Tan, Changbao Wang, Buyu Li, Quanquan Li, Wanli Ouyang, Changqing Yin,
  and Junjie Yan.
\newblock Equalization loss for long-tailed object recognition.
\newblock In {\em Proceedings of the IEEE/CVF Conference on Computer Vision and
  Pattern Recognition(CVPR)}, 2020.

\bibitem{tang2020long}
Kaihua Tang, Jianqiang Huang, and Hanwang Zhang.
\newblock Long-tailed classification by keeping the good and removing the bad
  momentum causal effect.
\newblock In {\em Advances in Neural Information Processing Systems(NeurIPS)},
  2020.

\bibitem{van2018inaturalist}
Grant Van~Horn, Oisin Mac~Aodha, Yang Song, Yin Cui, Chen Sun, Alex Shepard,
  Hartwig Adam, Pietro Perona, and Serge Belongie.
\newblock The inaturalist species classification and detection dataset.
\newblock In {\em Proceedings of the IEEE/CVF Conference on Computer Vision and
  Pattern Recognition(CVPR)}, 2018.

\bibitem{wang2019data}
Hao Wang, Qilong Wang, Fan Yang, Weiqi Zhang, and Wangmeng Zuo.
\newblock Data augmentation for object detection via progressive and selective
  instance-switching, 2019.

\bibitem{wang2020devil}
Tao Wang, Yu Li, Bingyi Kang, Junnan Li, Junhao Liew, Sheng Tang, Steven Hoi,
  and Jiashi Feng.
\newblock The devil is in classification: A simple framework for long-tail
  instance segmentation.
\newblock In {\em Proceedings of the European Conference on Computer
  Vision(ECCV)}, 2020.

\bibitem{wang2020few}
Xin Wang, Thomas~E. Huang, Trevor Darrell, Joseph~E Gonzalez, and Fisher Yu.
\newblock Frustratingly simple few-shot object detection.
\newblock In {\em International Conference on Machine Learning(ICML)}, 2020.

\bibitem{wang2018low}
Yu-Xiong Wang, Ross Girshick, Martial Hebert, and Bharath Hariharan.
\newblock Low-shot learning from imaginary data.
\newblock In {\em Proceedings of the IEEE/CVF Conference on Computer Vision and
  Pattern Recognition(CVPR)}, 2018.

\bibitem{wang2017learning}
Yu-Xiong Wang, Deva Ramanan, and Martial Hebert.
\newblock Learning to model the tail.
\newblock In {\em Advances in Neural Information Processing Systems(NeurIPS)},
  2017.

\bibitem{wu2020self}
Jialian Wu, Chunluan Zhou, Qian Zhang, Ming Yang, and Junsong Yuan.
\newblock Self-mimic learning for small-scale pedestrian detection.
\newblock In {\em Proceedings of the 28th ACM International Conference on
  Multimedia(ACM MM)}, 2020.

\bibitem{wu2020solving}
Tz-Ying Wu and Pedro Morgado.
\newblock Solving long-tailed recognition with deep realistic taxonomic
  classi{\"\i}$\neg$ er.
\newblock In {\em European Conference on Computer Vision(ECCV)}, 2020.

\bibitem{wu2019detectron2}
Yuxin Wu, Alexander Kirillov, Francisco Massa, Wan-Yen Lo, and Ross Girshick.
\newblock Detectron2.
\newblock \url{https://github.com/facebookresearch/detectron2}, 2019.

\bibitem{zhang2018context}
Hang Zhang, Kristin Dana, Jianping Shi, Zhongyue Zhang, Xiaogang Wang, Ambrish
  Tyagi, and Amit Agrawal.
\newblock Context encoding for semantic segmentation.
\newblock In {\em Proceedings of the IEEE/CVF Conference on Computer Vision and
  Pattern Recognition(CVPR)}, 2018.

\bibitem{zhang2020resnest}
Hang Zhang, Chongruo Wu, Zhongyue Zhang, Yi Zhu, Zhi Zhang, Haibin Lin, Yue
  Sun, Tong He, Jonas Muller, R. Manmatha, Mu Li, and Alexander Smola.
\newblock Resnest: Split-attention networks.
\newblock {\em arXiv preprint}, 2020.

\bibitem{zhong2019unequal}
Yaoyao Zhong, Weihong Deng, Mei Wang, Jiani Hu, Jianteng Peng, Xunqiang Tao,
  and Yaohai Huang.
\newblock Unequal-training for deep face recognition with long-tailed noisy
  data.
\newblock In {\em Proceedings of the IEEE Conference on Computer Vision and
  Pattern Recognition(CVPR)}, 2019.

\bibitem{zhou2020bbn}
Boyan Zhou, Quan Cui, Xiu-Shen Wei, and Zhao-Min Chen.
\newblock Bbn: Bilateral-branch network with cumulative learning for
  long-tailed visual recognition.
\newblock In {\em Proceedings of the IEEE/CVF Conference on Computer Vision and
  Pattern Recognition(CVPR)}, 2020.

\bibitem{zhou2017places}
Bolei Zhou, Agata Lapedriza, Aditya Khosla, Aude Oliva, and Antonio Torralba.
\newblock Places: A 10 million image database for scene recognition.
\newblock {\em IEEE transactions on pattern analysis and machine
  intelligence(TPAMI)}, 2017.

\bibitem{zhou2017scene}
Bolei Zhou, Hang Zhao, Xavier Puig, Sanja Fidler, Adela Barriuso, and Antonio
  Torralba.
\newblock Scene parsing through ade20k dataset.
\newblock In {\em Proceedings of the IEEE/CVF Conference on Computer Vision and
  Pattern Recognition(CVPR)}, 2017.

\bibitem{zhu2020cvpods}
Benjin Zhu*, Feng Wang*, Jianfeng Wang, Siwei Yang, Jianhu Chen, and Zeming Li.
\newblock cvpods: All-in-one toolbox for computer vision research, 2020.

\bibitem{zhu2014capturing}
Xiangxin Zhu, Dragomir Anguelov, and Deva Ramanan.
\newblock Capturing long-tail distributions of object subcategories.
\newblock In {\em Proceedings of the IEEE/CVF Conference on Computer Vision and
  Pattern Recognition(CVPR)}, 2014.

\end{thebibliography}
}
\appendix

\begin{table*}[t]
	\centering
	\resizebox{0.8\textwidth}{!}{
		\begin{tabular}{c|l|c|ccc|c|ccc}
			\toprule
			\multirow{2}{*}{\textbf{Backbone}}  & \multirow{2}{*}{\textbf{Method}} & \multicolumn{4}{c|}{\textbf{ResNet}} & \multicolumn{4}{c}{\textbf{ResNeXt}}  \\
			\cmidrule{3-10}
			&             &\textbf{Average} & \textbf{Many } & \textbf{Medium  } & \textbf{Few  } &\textbf{Average} & \textbf{Many  } & \textbf{Medium  } & \textbf{Few  } \\
			\midrule
			\multirow{4}{*}{*-50}            &Baseline    & 41.6	           & 64.0           &    33.8         &  5.8           
														  & 44.4	          & 65.9                  &       37.5     &  7.7 \\
						  &Baseline$^*$                   &  48.4  & 68.4   &  41.7    & 15.2                 
						                                  &  49.2   & 68.9   &  42.8  & 15.6\\
						                                  												 \cmidrule{2-10}
			& \textbf{DisAlign}                           &  51.3   &  59.9  &  49.9   &  31.8      
														  &  52.6   &  61.5  &  50.7   &  33.1\\
			& \textbf{DisAlign$^{*}$}                     &  52.9   &  61.3  &  52.2   &  31.4
														  &  53.4  &   62.7  &  52.1   &  31.4\\
			\midrule
			\multirow{4}{*}{*-101}            &Baseline    & 44.2	           & 66.6           &    36.8         &  7.1           
														   & 44.8	          & 66.2            &    37.8         &  8.6 \\
											  &Baseline$^*$  &   49.5 &    69.3 &  43.1 &  15.9                
											  				 &    50.0 &   69.9 &  43.7 &  15.9 \\
											 												 \cmidrule{2-10}
			& \textbf{DisAlign}                       &   52.7 &  61.7 &  51.1&  32.4    
													  &   53.6 &    63.3 &     51.2 &  34.6 \\
			& \textbf{DisAlign$^{*}$}                 &   54.1 &  63.2 &  53.1 &   31.9
													  &   54.6 &  64.7 &  53.0 &   31.7 \\
			\midrule
			\multirow{4}{*}{*-152}            &Baseline    & 44.9  & 66.9   &   37.7 &  7.7   
														   & 47.8  & 69.1   &   41.4   &  10.4 \\
						  &Baseline$^*$                   &  50.2 &  70.1  &       43.9 &     16.1
						  								  &   50.5  &  70.0 &  44.4 & 16.5\\
						  												 \cmidrule{2-10}
			& \textbf{DisAlign}                           &   53.7 &    62.8 &        51.9 &    34.2         
														  &   54.5 &  64.5 &      52.0 &     34.7  \\
			& \textbf{DisAlign$^{*}$}                     &   54.8 &    63.9 &        53.9 &  32.5      
			                                              &   55.0 &   65.1 &      53.3 &    32.2 \\
			\bottomrule
	\end{tabular}} 
		\caption{\textbf{Top-1 Accuracy on ImageNet-LT test set.} All models use the feature extractor and original classifier head trained with 90 epoch in joint learning stage, $*$ denotes the model uses cosine classifier head.}
	\label{tab:imagenet_lt}
\end{table*}

\section{Experiments of Image Classification}

In this section, we first introduce the dataset and evaluation metrics for image classification task in Sec.\ref{cls:dataset}.
Then the training configuration will be detailed in Sec.\ref{cls:training}, followed by results on three benchmarks in Sec.\ref{cls:results}. 

\subsection{Dataset and Evaluation Metrics}\label{cls:dataset}
\paragraph{Datasets} 
To demonstrate our methods, we conduct experiments on three large-scale long-tailed datasets, including Places-LT\cite{liu2019large}, ImageNet-LT\cite{liu2019large}, and iNaturalist 2018\cite{van2018inaturalist}. Places-LT and ImageNet-LT are artificially generated by sampling a subset from their balanced versions (Places-365\cite{liu2019large} and ImageNet-2012\cite{deng2009imagenet}) following the \textit{Parento distribution}. 
iNaturalist 2018 is a real-world, naturally long-tailed dataset, consisting of samples from 8,142 species. 

\paragraph{Evaluation Metrics}

We report the class-balanced average \textit{Top-1} accuracy on the corresponding validation/test set, and also calculate the accuracy of three disjoint subsets, `Many', `Medium' and `Few', which are defined according to the amount of training data per class~\cite{kang2019decoupling}.

\subsection{Training Configuration}\label{cls:training}

\paragraph{Configuration Detail}
Following \cite{kang2019decoupling}, we use PyTorch\cite{paszke2019pytorch} framework for all experiments.
For \textit{ImageNet-LT}, we report performance with ResNet-\{50,101,152\} and ResNeXt-\{50,101,152\} and mainly use ResNet-50 for ablation study. 
For \textit{iNaturalist 2018}, performance is reported with ResNet-\{50,101,152\}. 
For \textit{Places-LT}, ResNet-152 is used as backbone and we pre-train it on the full ImageNet-2012 dataset. 

We use the SGD optimizer with momentum 0.9, batch size 256, cosine learning rate schedule gradually decaying from 0.1 to 0, and image resolution 224$\times$224. For the joint learning stage, the backbone network and original classifier head are jointly trained with 90 epochs for ImageNet-LT, and 90/200 epochs for iNaturalist-2018. For the Places-LT dataset, the models are trained with 30 epochs with the all layers frozen expect the last ResNet block in the first stage.

\paragraph{Implementation of Our Method}

In the second distribution alignment stage, we restart the learning rate and train it for 10/30 epochs as \cite{kang2019decoupling} while keeping the backbone network and original classifier head fixed(10 epochs for ImageNet-LT and Places-LT, 30 epochs for iNaturalist-2018). For all three datasets, we set the generalized re-weight scale $\rho=1.2$ for dot-product classifier head, $\rho=1.5$ for cosine normalized classifier head. The $\alpha$ and $\beta$ are initialized with 1.0 and 0.0, respectively.

\subsection{Detailed Experimental Results}\label{cls:results}
\paragraph{ImageNet-LT.} We present the detailed quantitative results for ImageNet-LT in Table \ref{tab:imagenet_lt}. 

\paragraph{iNaturalist and Places-LT.}
To further demonstrate our method, we conduct experiments on two extra large-scale long-tail benchmarks and report the performance in Table \ref{tab:inaturalist} and Table \ref{tab:places}.

\subsection{Ablation Study}

\begin{table}[!h]
	\centering
	\resizebox{0.8\linewidth}{!}{
		\begin{tabular}{c|c|c|c}
			\toprule
			\textbf{Method} & \textbf{Calibration} & \textbf{G-RW} & \textbf{Top-1 Acc} \\
			\midrule
			Baseline$^*$ & -  & - &  49.2  \\
			\midrule
			cRT$^*$  & \xmark   &  \xmark   & 49.7   \\
			-  & \xmark  &  \cmark  & 51.9  \\
			\midrule
			DisAlign$^*$ &  \cmark &  \cmark  & 53.4 \\
			\midrule
	\end{tabular}}
	\caption{\textbf{Influence of Model Components.} Backbone is ResNeXt-50, $*$ means cosine classifier.}
	\label{tab:ablation_1}
\end{table}

\paragraph{Influence of Model Components}
We report an ablation study of the two main components of our method with ResNeXt-50 in Tab.~\ref{tab:ablation_1},
which shows that both adaptive calibration and generalized re-weighting(G-RW) contribute to the performance improvement of our approach.

\paragraph{Analysis of the Calibration}
We plot the learned magnitude and margin according to the class sizes below. They share a similar trend, in which the tail/body classes have larger value than head. Thus our calibration alleviates the bias in the original prediction by boosting the tail scores.

\begin{figure*}[ht]
    \centering
    \includegraphics[width=0.85\linewidth]{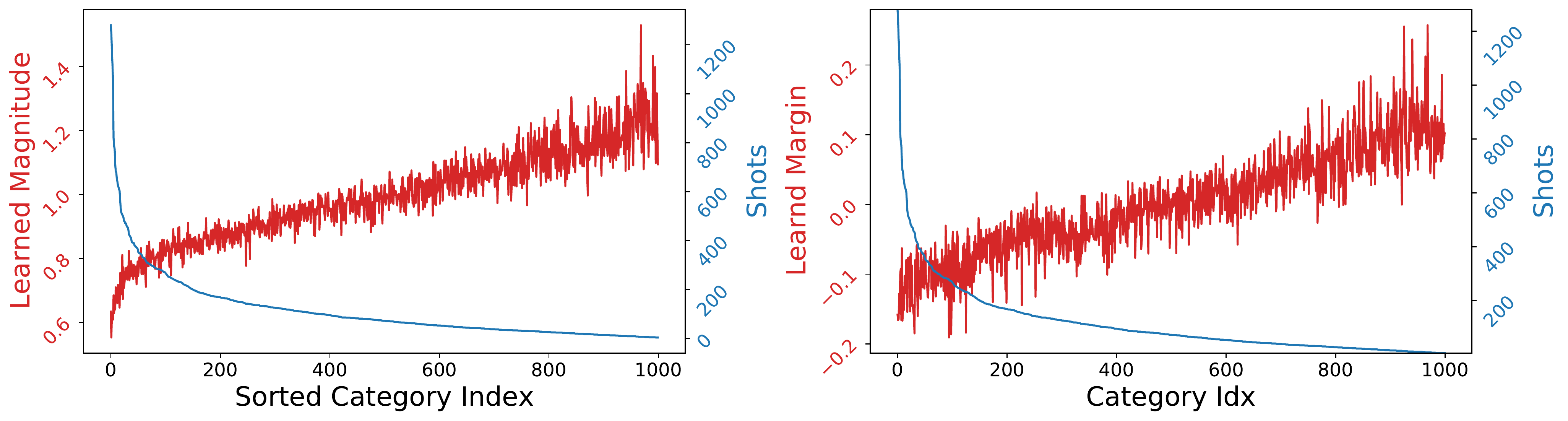}
    \caption{\textbf{Analysis of the Calibration.} We use model trained on ImageNet-LT with ResNeXt-50 for analysis.}
    \label{fig:ad}
\end{figure*}

\paragraph{Confidence Score}
We study confidence-based calibration in the table below, which 
shows that the input-aware calibration outperforms the input-agnostic counterpart and the baselines using only magnitude or margin. We also observe that the example whose biased prediction probability is low on its ground-truth class tends to be improved with higher confidence.
\begin{table}[ht]
	\centering
	\resizebox{0.99\linewidth}{!}{
		\begin{tabular}{l|c|cc|cc|cc}
			\toprule
				\textbf{Generalized Re-weighting}  &   &   \cmark    &     \cmark    &   \cmark   &     \cmark  &     \cmark &  \cmark  \\
				\midrule
					 \textbf{Magnitude\small{(w/o Confidence)}} &   &  \cmark    &   &  & &\cmark      \\
				\textbf{Magnitude}&    &  & \cmark   & &             & &\cmark        \\
				\midrule
				\textbf{Margin\small{(w/o Confidence)}} &    &  &    & \cmark &  & \cmark          \\
				 \textbf{Margin}         &    &  &  &    &   \cmark  & &   \cmark           \\
				\midrule
				 \textbf{Average Accuracy}    &  41.6 & 49.9 &50.1 &    49.6 & 49.9 & 51.0 & 51.3 \\
			\bottomrule
		\end{tabular}
	}
	\caption{\textbf{Ablation of the Confidence Score.} We extend the Tab.5(main paper) to analyze the influence of confidence score.}
\end{table}

\begin{table*}[htbp]
	\centering
	\resizebox{0.8\textwidth}{!}{
		\begin{tabular}{c|l|c|ccc|c|ccc}
			\toprule
			\multirow{2}{*}{\textbf{Backbone}}  & \multirow{2}{*}{\textbf{Method}} & \multicolumn{4}{c|}{\textbf{90 Epoch}} & \multicolumn{4}{c}{\textbf{200 Epoch}}  \\
			\cmidrule{3-10}
			&             &\textbf{Average} & \textbf{Many } & \textbf{Medium  } & \textbf{Few  } &\textbf{Average} & \textbf{Many  } & \textbf{Medium  } & \textbf{Few  } \\
			\midrule
			\multirow{4}{*}{ResNet-50}           & Baseline   & 61.7  & 72.2 & 63.0 & 57.2 
															   & 65.8 & 75.7 & 66.9 & 61.7 \\
												 & Baseline$^*$  &   64.8 &     75.8 &       66.6&    59.7 
												 				 &   66.2 &     77.3 &       68.3&    60.7\\
												 \cmidrule{2-10}
												 & \textbf{DisAlign} &  67.8 & 64.1 & 68.5 & 67.9
												 				     &  70.6 &     69.0 &     71.1 &   70.2  \\
												 & \textbf{DisAlign$^{*}$} &   69.5 &  61.6 &  70.8 &  69.9 
												 							&  70.2 & 68.0 &   71.3 &   69.4   \\
			\midrule
			\multirow{4}{*}{ResNet-101}          & Baseline    & 64.6 & 75.9 & 66.0 & 59.9
															   & 67.3 & 75.5 & 68.9 & 63.2\\
												 & Baseline$^*$   & 66.4 &   76.8& 68.5&   61.1 
												 			   &    68.0 &     78.9&  69.7&  63.0 \\
																								 \cmidrule{2-10}
												 & \textbf{DisAlign}  &   70.0 & 68.3 &     70.4 & 69.9  
												 						&  72.9 &     73.0 &    73.5 &   72.1\\
												 & \textbf{DisAlign$^{*}$}  &  70.8 &  65.4 &   72.2&   70.4 
												 						&   71.9 &  69.3 &      72.6 &     71.8\\
			\midrule
			\multirow{4}{*}{ResNet-152}          & Baseline     & 65.0 & 75.2 & 66.3 & 60.7
																& 69.0 & 78.2 & 70.6 & 64.7\\
												 & Baseline$^*$   &  67.3 & 77.8 & 69.4 & 61.8
												 				 &   69.0 & 78.5 &  71.0 & 64.0 \\
												 												 \cmidrule{2-10}
												 & \textbf{DisAlign}   &   71.3 & 70.7 &     71.8 &  70.8 
												 				    	&   74.1 & 74.9 &  74.4 & 73.5  \\
												 & \textbf{DisAlign$^{*}$}  &   71.7 &      67.1 & 73.0 & 71.3
												 							&    72.8&  70.6 & 73.6 & 72.3\\
			\bottomrule
	\end{tabular}} 
		\caption{Top-1 Accuracy on iNaturalist 2018 with different backbones(ResNet-\{50,101,152\}) and different training epochs(90 \& 200), $*$ denotes the model uses cosine classifier head.}
	\label{tab:inaturalist}
\end{table*}

\begin{table}[!tbp]
	\centering
	\resizebox{0.48\textwidth}{!}{
		\begin{tabular}{l|l|c|ccc}
			\toprule
			\multirow{2}{*}{\textbf{Backbone}}  & \multirow{2}{*}{\textbf{Method}} & \multicolumn{4}{c}{\textbf{Top-1 Accuracy}}   \\
			\cmidrule{3-6}
			&              &\textbf{Average} & \textbf{Many  } & \textbf{Medium  } & \textbf{Few  } \\
			\midrule
			\multirow{2}{*}{R-50}                & Baseline    &   29.2 & 45.3 &  25.5 &  8.0  \\
												 & \textbf{DisAlign} &  37.8 & 39.3 & 40.7 & 28.5\\
			\midrule
			\multirow{2}{*}{R-101}               & Baseline    &  30.2 & 46.1 &   26.9 &   8.4 \\
												 & \textbf{DisAlign} &    38.5 &   39.1 &  42.0 & 29.1 \\
			\midrule
			\multirow{2}{*}{R-152}          & Baseline     & 30.2 & 45.7 & 27.3 & 8.2  \\
												 & \textbf{DisAlign} & 39.3 & 40.4 & 42.4 & 30.1   \\
			\bottomrule
	\end{tabular}} 
		\caption{Top-1 Accuracy on Places-LT with different backbones(ResNet-\{50,101,152\}).}
	\label{tab:places}
\end{table}

\section{Experiments of Semantic Segmentation}
Similar to image classification, the large-scale semantic segmentation task still suffers from the long-tail data distribution. To further validate the effectiveness of our method, we also apply DisAlign on large-scale semantic segmentation benchmark: ADE-20k. 

\subsection{Dataset and Evaluation}
\paragraph{Dataset.} ADE20K dataset is a scene parsing benchmark, which contains 150 stuff/object categories. The dataset includes 20K/2K/3K images for training, validation, and testing. Compared with the image classification\cite{liu2019large}, the imbalance of ADE20K is more serve than the image classification, which has an imbalance ratio of \textbf{788}(Max/Min). Follow the similar protocol in image classification, we divide the 150 categories into 3 groups according to the ratio of pixel number over the whole dataset. Specifically, three disjoint subsets are: \textit{head classes}(classes each with a ratio over 1.0\%), \textit{body classes}(classes each with a ratio ranging from 0.1\% to 1\%) and \textit{tail classes}(classes under a ratio of 0.1\%), the complete list of the split is reported in Tab.\ref{tab:group}.

\paragraph{Evaluation.}  For the evaluation metric, we use the mean intersection of union(mIoU) and mean pixel accuracy(mAcc). We also report the mIoU and mAcc of each group(head, body and tail) for clarity.

\begin{table*}[h!]

	\centering
	 \resizebox{0.999\textwidth}{!}{
	\begin{tabular}{l|l|c||c|l|lll||l|lll}
	\toprule
	\multirow{2}{*}{\textbf{Framework}} & \multirow{2}{*}{\textbf{B}}  & \multirow{2}{*}{\textbf{Method}} & \multirow{2}{*}{\textbf{Aug}} & \multicolumn{4}{c||}{\textbf{Mean IoU}} & \multicolumn{4}{c}{\textbf{Mean Accuracy}}\\
	\cmidrule{5-12}
				 & & & & \textbf{Average} & \textbf{Head} & \textbf{Body} & \textbf{Tail} &  \textbf{Average} & \textbf{Head} & \textbf{Body} & \textbf{Tail} \\
	\midrule
	\multirow{12}{*}{FCN\cite{shelhamer2017fully}}  
			&\multirow{4}{*}{R-50} & Baseline & \xmark    & 36.1  &  62.5   &  38.1   & 27.6 
														  & 45.4  &  76.9   &  48.8   & 34.5 \\
		   & & \textbf{DisAlign} & \xmark & 37.5\more{(+1.4)}  & 62.6\more{(+0.1)}  & 40.2\more{(+2.1)} & 28.8\more{(+1.2)} 
										 & 49.9\more{(+4.5)}  & 76.7\small{(-0.2)} & 54.9\more{(+6.1)} & 39.0\more{(+4.5)}\\
										 \cmidrule{3-12}
					    &  & Baseline & \cmark & 38.1  &  64.6   &  40.0   & 29.6 
																  & 46.3  &  78.6   &  49.3   & 35.4 \\
	      &  &  \textbf{DisAlign} & \cmark & 40.1\more{(+2.0)}  & 65.0\more{(+0.4)}  & 42.8\more{(+2.8)} & 31.3\more{(+1.7)}   
										   & 51.4\more{(+5.1)}  & 78.6\small{(+0.0)} & 56.1\more{(+6.8)} & 40.6\more{(+5.2)}\\
	\cmidrule{2-12}
	     &  \multirow{4}{*}{R-101} & Baseline & \xmark &  39.9  &  65.3   &  42.0   & 31.7 
														&  49.6  &  79.1   &  52.6   & 39.6 \\
	 &  &\textbf{DisAlign}  &\xmark &  41.8\more{(+1.9)}  &  65.5\more{(+0.2)}   & 44.1\more{(+2.1)}  & 33.7\more{(+2.0)}  
									&  54.7\more{(+5.1)}  &  79.0\small{(-0.1)}  & 58.6\more{(+6.0)}  & 45.2\more{(+5.6)}\\
									\cmidrule{3-12}
	     &                         & Baseline  & \cmark &  41.4  &   67.0  &  43.3   & 33.2 
													  &  50.2  &   80.6  &  52.9   & 40.1 \\
	     & &\textbf{DisAlign}   & \cmark &  43.7\more{(+2.3)}  & 67.4\more{(+0.4)}  & 46.1\more{(+2.8)} & 35.7\more{(+2.5)}   
								  &  55.9\more{(+5.7)}  & 80.6\small{(+0.0)} & 59.7\more{(+6.8)} & 46.4\more{(+6.3)}  \\
	\cmidrule{2-12}
	 & \multirow{4}{*}{S-101} &  Baseline & \xmark  & 45.6  &  66.6  &  47.5   & 38.6 
											 & 57.8  &  78.8  &  62.1   & 48.9 \\
	  &  &\textbf{DisAlign}   & \xmark  & 46.2\more{(+0.6)}  & 66.6\small{(+0.0)}   & 48.0\more{(+0.4)}    & 39.4\more{(+0.8)}   
								& 60.3\more{(+2.5)}  & 79.1\more{(+0.3)}& 64.9\more{(+2.8)} & 51.7\more{(+2.8)}\\
								\cmidrule{3-12}
	  &  &  Baseline    & \cmark  & 46.2  & 67.6   & 48.0  & 39.1 
								 & 57.3  & 79.4   & 61.7  & 48.2   \\
	    & &\textbf{DisAlign}  & \cmark     & 46.9\more{(+0.7)}  & 67.7\more{(+0.1)}  & 48.2\more{(+0.2)}  & 40.3\more{(+1.2)}
							  & 60.1\more{(+2.8)} &  79.7\more{(+0.3)}  & 64.2\more{(+2.5)}  & 51.9\more{(+3.7)} \\
	\midrule
	
	\multirow{12}{*}{DeepLabV3$^{+}$\cite{deeplabv3plus2018} }
		&\multirow{4}{*}{R-50} & Baseline & \xmark  &  43.9  &  66.6  &   47.1  &   35.6 
													  &  54.9  &  79.4  &   60.3  &   44.5     \\
	     & & \textbf{DisAlign}    & \xmark &  44.4\more{(+0.5)} & 66.6\small{(+0.0)} & 47.2\more{(+0.1)} & 36.5\more{(+0.9)} 
									     &  57.2\more{(+2.3)} & 79.8\more{(+0.4)}  & 62.3\more{(+2.0)} & 47.5\more{(+3.0)} \\
																			  \cmidrule{3-12}
	 &        & Baseline  & \cmark &  44.9  &  67.7  &   48.3  &   36.4 
						 	  &  55.0  &  80.1  &   60.8  &   44.1 \\
	      &  & \textbf{DisAlign}   & \cmark &  45.7\more{(+0.8)} & 67.7\small{(+0.0)} & 48.6\more{(+0.3)} & 37.8\more{(+1.4)}
									 &  57.3\more{(+2.3)} & 80.8\more{(+0.7)} & 63.0\more{(+2.2)} & 46.9\more{(+2.8)} \\
	  \cmidrule{2-12}
	 &\multirow{4}{*}{R-101} & Baseline & \xmark &  45.5 & 67.6 & 48.2 & 37.6 & 56.4 & 80.1 & 61.2 & 46.6     \\
	      &  & \textbf{DisAlign}  & \xmark &  46.0\more{(+0.5)} & 67.6\small{(+0.0)} & 48.4\more{(+0.2)} & 38.5\more{(+0.9)} 
								  &  59.1\more{(+2.7)} & 80.5\more{(+0.4)} & 63.8\more{(+2.6)} & 49.9\more{(+3.3)} \\
		  \cmidrule{3-12}
	  &  & Baseline & \cmark &  46.4 & 68.7  & 49.0 & 38.4 & 56.7 & 80.9 & 61.5 & 46.7 \\
      &  & \textbf{DisAlign}     & \cmark &  47.1\more{(+0.7)} & 68.7\small{(+0.0)} & 49.4\more{(+0.4)} & 39.6\more{(+1.2)}
							  &  59.5\more{(+2.8)} & 81.4\more{(+0.5)} & 64.2\more{(+2.7)} & 50.3\more{(+3.6)} \\
		   \cmidrule{2-12}
	 &\multirow{4}{*}{S-101}  & Baseline &  \xmark   &  46.5  & 68.0  & 49.1 & 38.8 
								 	&  58.1  & 80.1  & 63.4 & 48.5     \\
	     & & \textbf{DisAlign}  & \xmark   &  46.9\more{(+0.4)} & 67.8\small{(-0.2)} & 49.2\more{(+0.1)} & 39.6\more{(+0.8)} 
							 &  60.7\more{(+2.6)} & 80.5\more{(+0.4)}  & 65.5\more{(+2.1)}  & 51.9\more{(+3.4)}\\
	\cmidrule{3-12}
	  &     & Baseline             & \cmark &  47.3 & 69.0 & 49.7 & 39.7
								  &  58.1 & 80.8 & 63.4 & 48.2\\
	   &  & \textbf{DisAlign}      & \cmark &  47.8\more{(+0.5)} & 68.9\small{(-0.1)} & 49.8\more{(+0.1)} & 40.7\more{(+1.0)} 
								   &  60.1\more{(+2.0)} & 81.0\more{(+0.2)}  & 65.5\more{(+2.1)} & 52.0\more{(+3.8)}\\
	\bottomrule
	\end{tabular}}
	\caption{\textbf{Results on ADE-20K:} All baseline models are trained with a image size of 512x512 and 160K iteration in total. \textbf{Aug} denotes multi-scale is used for inference.}
	\label{tab:ade20k}
\end{table*}

\begin{table*}[]
	\centering
	 \resizebox{0.999\textwidth}{!}{
	\begin{tabular}{lll|lll|lll|lll|lll}
		\toprule
	\textbf{Category}   & \textbf{Ratio}  & \textbf{Group} & \textbf{Category}   & \textbf{Ratio} &  \textbf{Group}  & \textbf{Category}     & \textbf{Ratio}    &  \textbf{Group}  &\textbf{Category}    & \textbf{Ratio}   & \textbf{Group}  &\textbf{Category}    & \textbf{Ratio}    &  \textbf{Group}  \\
	\midrule
	'wall'        & 0.1576, & Head  & 'armchair'          & 0.0044, & Body  & 'river'               & 0.0015, & Body  & 'airplane'       & 0.0007, & Tail  & 'food'            & 0.0005, & Tail  \\
	'building'   & 0.1072, & Head  & 'seat'             & 0.0044, & Body  & 'bridge'             & 0.0015, & Body  & 'dirt track'    & 0.0007, & Tail  & 'step'           & 0.0004, & Tail  \\
	'sky'        & 0.0878, & Head  & 'fence'            & 0.0033, & Body  & 'bookcase'           & 0.0014, & Body  & 'apparel'       & 0.0007, & Tail  & 'tank'           & 0.0004, & Tail  \\
	'floor'      & 0.0621, & Head  & 'desk'             & 0.0031, & Body  & 'blind'              & 0.0014, & Body  & 'pole'          & 0.0006, & Tail  & 'trade name'     & 0.0004, & Tail  \\
	'tree'       & 0.048,  & Head  & 'rock'             & 0.003,  & Body  & 'coffee table'       & 0.0014, & Body  & 'land'          & 0.0006, & Tail  & 'microwave'      & 0.0004, & Tail  \\
	'ceiling'    & 0.045,  & Head  & 'wardrobe'         & 0.0027, & Body  & 'toilet'             & 0.0014, & Body  & 'bannister'     & 0.0006, & Tail  & 'pot'            & 0.0004, & Tail  \\
	'road'       & 0.0398, & Head  & 'lamp'             & 0.0026, & Body  & 'flower'             & 0.0014, & Body  & 'escalator'     & 0.0006, & Tail  & 'animal'         & 0.0004, & Tail  \\
	'bed'        & 0.0231, & Head  & 'bathtub'          & 0.0024, & Body  & 'book'               & 0.0013, & Body  & 'ottoman'       & 0.0006, & Tail  & 'bicycle'        & 0.0004, & Tail  \\
	'windowpane' & 0.0198, & Head  & 'railing'          & 0.0024, & Body  & 'hill'               & 0.0013, & Body  & 'bottle'        & 0.0006, & Tail  & 'lake'           & 0.0004, & Tail  \\
	'grass'      & 0.0183, & Head  & 'cushion'          & 0.0023, & Body  & 'bench'              & 0.0013, & Body  & 'buffet'        & 0.0006, & Tail  & 'dishwasher'     & 0.0004, & Tail  \\
	'cabinet'    & 0.0181, & Head  & 'base'             & 0.0023, & Body  & 'countertop'         & 0.0012, & Body  & 'poster'        & 0.0006, & Tail  & 'screen'         & 0.0004, & Tail  \\
	'sidewalk'   & 0.0166, & Head  & 'box'              & 0.0022, & Body  & 'stove'              & 0.0012, & Body  & 'stage'         & 0.0006, & Tail  & 'blanket'        & 0.0004, & Tail  \\
	'person'     & 0.016,  & Head  & 'column'           & 0.0022, & Body  & 'palm'               & 0.0012, & Body  & 'van'           & 0.0006, & Tail  & 'sculpture'      & 0.0004, & Tail  \\
	'earth'      & 0.0151, & Head  & 'signboard'        & 0.002,  & Body  & 'kitchen island'     & 0.0012, & Body  & 'ship'          & 0.0006, & Tail  & 'hood'           & 0.0004, & Tail  \\
	'door'       & 0.0118, & Head  & 'chest of drawers' & 0.0019, & Body  & 'computer'           & 0.0011, & Body  & 'fountain'      & 0.0005, & Tail  & 'sconce'         & 0.0003, & Tail  \\
	'table'      & 0.011,  & Head  & 'counter'          & 0.0019, & Body  & 'swivel chair'       & 0.001,  & Tail  & 'conveyer belt' & 0.0005, & Tail  & 'vase'           & 0.0003, & Tail  \\
	'mountain'   & 0.0109, & Head  & 'sand'             & 0.0018, & Body  & 'boat'               & 0.0009, & Tail  & 'canopy'        & 0.0005, & Tail  & 'traffic light'  & 0.0003, & Tail  \\
	'plant'      & 0.0104, & Head  & 'sink'             & 0.0018, & Body  & 'bar'                & 0.0009, & Tail  & 'washer'        & 0.0005, & Tail  & 'tray'           & 0.0003, & Tail  \\
	'curtain'    & 0.0104, & Head  & 'skyscraper'       & 0.0018, & Body  & 'arcade machine'     & 0.0009, & Tail  & 'plaything'     & 0.0005, & Tail  & 'ashcan'         & 0.0003, & Tail  \\
	'chair'      & 0.0103, & Head  & 'fireplace'        & 0.0018, & Body  & 'hovel'              & 0.0009, & Tail  & 'swimming pool' & 0.0005, & Tail  & 'fan'            & 0.0003, & Tail  \\
	'car'        & 0.0098, & Body  & 'refrigerator'     & 0.0018, & Body  & 'bus'                & 0.0009, & Tail  & 'stool'         & 0.0005, & Tail  & 'pier'           & 0.0003, & Tail  \\
	'water'      & 0.0074, & Body  & 'grandstand'       & 0.0018, & Body  & 'towel'              & 0.0008, & Tail  & 'barrel'        & 0.0005, & Tail  & 'crt screen'     & 0.0003, & Tail  \\
	'painting'   & 0.0067, & Body  & 'path'             & 0.0018, & Body  & 'light'              & 0.0008, & Tail  & 'basket'        & 0.0005, & Tail  & 'plate'          & 0.0003, & Tail  \\
	'sofa'       & 0.0065, & Body  & 'stairs'           & 0.0017, & Body  & 'truck'              & 0.0008, & Tail  & 'waterfall'     & 0.0005, & Tail  & 'monitor'        & 0.0003, & Tail  \\
	'shelf'      & 0.0061, & Body  & 'runway'           & 0.0017, & Body  & 'tower'              & 0.0008, & Tail  & 'tent'          & 0.0005, & Tail  & 'bulletin board' & 0.0003, & Tail  \\
	'house'      & 0.006,  & Body  & 'case'             & 0.0017, & Body  & 'chandelier'         & 0.0008, & Tail  & 'bag'           & 0.0005, & Tail  & 'shower'         & 0.0003, & Tail  \\
	'sea'        & 0.0053, & Body  & 'pool table'       & 0.0017, & Body  & 'awning'             & 0.0007, & Tail  & 'minibike'      & 0.0005, & Tail  & 'radiator'       & 0.0003, & Tail  \\
	'mirror'     & 0.0052, & Body  & 'pillow'           & 0.0017, & Body  & 'streetlight'        & 0.0007, & Tail  & 'cradle'        & 0.0005, & Tail  & 'glass'          & 0.0002, & Tail  \\
	'rug'        & 0.0046, & Body  & 'screen door'      & 0.0015, & Body  & 'booth'              & 0.0007, & Tail  & 'oven'          & 0.0005, & Tail  & 'clock'          & 0.0002, & Tail  \\
	'field'       & 0.004   & Body  & 'stairway'          & 0.0015  & Body  & 'television receiver' & 0.0007  & Tail  & 'ball'           & 0.0005  & Tail  & 'flag'            & 0.0002  & Tail \\	\bottomrule
	\end{tabular}}
	\caption{\textbf{Splits of ADE-20K:} The ratio of each category is reported according to \cite{zhou2017scene}.}
	\label{tab:group}
	\end{table*}

\begin{table*}[t]
	\centering
	\resizebox{0.8\textwidth}{!}{
	\begin{tabular}{c|l|c|ccc|c|ccc}
	\toprule
  
	  \multirow{2}{*}{\textbf{Backbone}} & \multirow{2}{*}{\textbf{Method}}  &   \multicolumn{4}{c|}{\textbf{BBox AP}} & \multicolumn{4}{c}{\textbf{Mask AP}} \\
	 \cmidrule{3-10}
	  &  &$\text{\textbf{AP}}_{bbox}$ & $\text{\textbf{AP}}_{bbox}^r$ & $\text{\textbf{AP}}_{bbox}^c$ & $\text{\textbf{AP}}_{bbox}^f$ & $\text{\textbf{AP}}_{mask}$& $\text{\textbf{AP}}_{mask}^r$ & $\text{\textbf{AP}}_{mask}^c$ & $\text{\textbf{AP}}_{mask}^f$ \\
	\midrule
	\multirow{4}{*}{ResNet-50 } & Baseline        & 	20.8  & 	3.3	  & 19.5  & 	29.4  & 	21.2  & 	3.7	  & 21.6  & 	28.4     \\
	& \textbf{DisAlgin}                   & 	23.9 &	7.5&	25.0 & 	29.1 &	24.2 &	8.5 &	26.2 &	28.0      \\
	 \cmidrule{2-10}
	& Baseline$^{*}$                    & 22.8 &	10.3 &	21.1 & 	30.1 &	23.8 &	11.5 &	23.7 &	28.9      \\
	& \textbf{DisAlgin$^{*}$}           &  25.6  &   13.7  &  25.6  &  30.5 &  26.3  & 14.9 &  27.6  &   29.2 \\
	\midrule
	\multirow{4}{*}{ResNet-101 } & Baseline       & 	22.2 &	2.6 & 21.1 &	31.6 & 22.6 &	2.7 &	22.8 & 	30.2     \\
	& \textbf{DisAlgin}                   & 	25.6 &	9.0 &	26.5 & 	30.9 &	25.8 &	10.3 & 	27.6	& 29.6\\
	 \cmidrule{2-10}
	& Baseline$^{*}$                          &24.5 &	10.1 &	23.2 &	31.8 &	25.1 &	11.2 &	25.2 &	30.4   \\
	& \textbf{DisAlgin$^{*}$}             & 27.5 &	15.9 &	27.6 &	32.0 &	28.2 &	17.8 &	29.7 &	30.5\\
	\midrule
	\multirow{4}{*}{ResNeXt-101 } & Baseline     &	24.5 &	3.9 &	24.1 &	33.1 &	25.0 & 	4.2 &	26.3 &	31.8    \\
	 & \textbf{DisAlgin}                  & 26.8 &	8.8 &	27.6 &	33.0 & 27.4 &	11.0 &	29.3 &	31.6 \\
	 \cmidrule{2-10}
	& Baseline$^{*}$                      & 26.9 &	12.1 &	26.1 &	33.8 &	27.7 &	15.2 &	28.2 &	32.2\\
	& \textbf{DisAlgin$^{*}$}             & 29.5 &	17.7 &	29.5 &	33.8 &	30.0 & 	19.6 &	31.5 &	32.3 \\
	\bottomrule
	\end{tabular}}
	\caption{\textbf{Results on LVIS v0.5 dataset with Mask R-CNN.} * denotes the model use cosine classifier head.}
	\label{tab:lvis_mask_rcnn}
	\end{table*}

\begin{table*}
	\centering
	\resizebox{0.8\textwidth}{!}{
	  \begin{tabular}{c|l|c|ccc|c|ccc}
		\toprule
	  
		  \multirow{2}{*}{\textbf{Backbone}} & \multirow{2}{*}{\textbf{Method}}  &   \multicolumn{4}{c|}{\textbf{BBox AP}} & \multicolumn{4}{c}{\textbf{Mask AP}} \\
		 \cmidrule{3-10}
		  &  &$\text{\textbf{AP}}_{bbox}$ & $\text{\textbf{AP}}_{bbox}^r$ & $\text{\textbf{AP}}_{bbox}^c$ & $\text{\textbf{AP}}_{bbox}^f$ & $\text{\textbf{AP}}_{mask}$& $\text{\textbf{AP}}_{mask}^r$ & $\text{\textbf{AP}}_{mask}^c$ & $\text{\textbf{AP}}_{mask}^f$ \\
	\midrule
	\multirow{4}{*}{ResNet-50 } & Baseline        &	25.2 & 3.7 & 24.3&	34.8 &	23.0 &	3.5 &	23.0 &	30.8     \\
	& \textbf{DisAlgin}                           & 28.7 &	9.0 & 30.2 &	34.6 & 26.1 &	8.4 &	28.1 &	30.7\\
	 \cmidrule{2-10}
	& Baseline$^{*}$                          &28.8 &	15.4 &	28.2 &	34.9 &	26.2 &	13.6 &	26.3 &	31.1   \\
	& \textbf{DisAlgin$^{*}$}             & 32.2 &	21.6 &	33.3 &	35.2 &	29.4 &	19.4 &	30.9 &	31.4\\
	\midrule
	\multirow{4}{*}{ResNet-101 } & Baseline       &26.1 & 3.4 &	25.4 &	35.9 &	24.0 &	3.3 &	24.2 &	32.0\\
	& \textbf{DisAlgin}                   & 	29.7 &8.1 &	31.7 &	35.8 &	27.3 &	7.8 &	29.7 &	32.0 \\
	 \cmidrule{2-10}
	& Baseline$^{*}$                          & 30.4 &	15.5 &	30.3 &	36.5 &	28.1 &	13.9 &	29.2 &	32.4        \\
	& \textbf{DisAlgin$^{*}$}             &   33.7 &  22.1 &  34.9 &  36.9 & 30.9 & 19.0 &  33.2 & 32.8\\
	\midrule
	\multirow{4}{*}{ResNeXt-101 } & Baseline    &28.4 &	4.6 &	28.6 &	37.5 &	26.1 &	4.6 &	27.2 &	33.4  \\
	 & \textbf{DisAlgin}                        & 31.3 &	9.5 &	33.2&	37.7 &	28.7 &	9.0 &	31.1 &	33.6\\
	 \cmidrule{2-10}
	& Baseline$^{*}$                      & 32.6 &	18.5 &	32.8 &	37.9 &	29.8 &	16.9 &	30.9 &	33.7\\
	& \textbf{DisAlgin$^{*}$}             & 34.7 & 24.6 & 35.3 & 38.1 &31.8 & 22.0 & 33.2 & 33.9 \\
	\bottomrule
	\end{tabular}}
	\caption{\textbf{Results on LVIS v0.5 dataset with Cascade R-CNN.} * denotes the model use cosine classifier head.}
	\label{tab:lvis_cascade_rcnn}
	\end{table*}
\subsection{Training Configuration}

We implement our method based on MMSegmentation toolkit\cite{mmsegmentation}. In the joint learning training phase, we set the learning rate to 0.01 initially, which gradually decreases to 0 by following the 'poly' strategy as \cite{zhang2018context}. The images are cropped to $512\times 512$ and augmented with randomly scaling(from 0.5 to 2.0) and flipping. ResNet-50, ResNet-101 and ResNeSt-101\cite{zhang2020resnest} are used as the backbone. For the evaluation metric, we use the mean intersection of union(mIoU) and mean pixel accuracy(mAcc). All models are trained with 160k iterations with a batch size of 32 based on 8 V100 GPUs. In the DisAlign stage, we follow a similar protocol as stage-1 and only training the model with 8k iterations. We set $\rho=0.3$ for all experiments.

\subsection{Quantitative Results}

We evaluate our method with two state-of-the-art segmentation models(FCN\cite{shelhamer2017fully} and DeepLabV3+\cite{deeplabv3plus2018})based on different backbone networks, ranging from ResNet-50, ResNet-101 to the latest ResNeSt-101, and report the performance in Tab.\ref{tab:ade20k}. 



\section{Experiments on LVIS Dataset}\label{sec:lvis}

\subsection{Dataset and Evaluation Protocol}
\paragraph{Dataset.} LVIS v0.5\cite{gupta2019lvis} dataset is a benchmark dataset for research on large vocabulary object detection and instance segmentation, which contains 56K images over 1230 categories for training, 5K images for validation. This challenging dataset is an appropriate benchmark to study the large-scale long-tail problem, where the categories can be binned into three types similar with ImageNet-LT: \textit{rare}(1-10 training images), \textit{common}(11-100 training images), and \textit{frequent}($>100$ training images).  
 \paragraph{Evaluation Protocol.} We evaluate our method on LVIS for object detection and instance segmentation. For evaluation, we use a COCO-style average precision(AP) metric that averages over categories and different box/mask intersection over union(IoU) threshold\cite{lin2014microsoft}. All standard LVIS evaluation metrics including $\text{AP}, \text{AP}^{r}, \text{AP}^{c}, \text{AP}^{f} $ for box bounding boxes and segmentation masks. Subscripts ‘r’, ‘c’, and ‘f’ refer to rare, common and frequent category subsets.
 

\subsection{Training Configuration}
\paragraph{Experimental Details.}
We train our models for object detection and instance segmentation based on Detecron2\cite{wu2019detectron2}, which is implemented in PyTorch. Unless specified, we use the ResNet backbone(pre-trained on ImageNet) with FPN\cite{lin2017feature}. 
Following the training procedure in \cite{gupta2019lvis}, we resize the images so that the shorter side is 800 pixels. All baseline experiments are conducted on 8 GPUs with 2 images per GPU for 90K iterations, with a learning rate of 0.02 which is decreased by 10 at the 60K and 80K iteration. We use SGD with a weight decay of 0.0001 and momentum of 0.9. Scale jitter is applied for all experiments in default same with \cite{gupta2019lvis}.

For the DisAlign, we freeze all network parameters and learn the magnitude and margin for extra 9K iterations with a learning rate of 0.02. Generalized re-weight is only used for fore-ground categories. Generalized re-weight scale $\rho$ is set to 0.8 for all experiments.

\subsection{Quantitative Results}
We report the detailed results in Table.\ref{tab:lvis_mask_rcnn} and Tab.\ref{tab:lvis_cascade_rcnn}.

\end{document}